# Echoes of Biases: How Stigmatizing Language Affects AI Performance

Yizhi Liu[1], Weiguang Wang[2], Guodong (Gordon) Gao[3], Ritu Agarwal[3]

[1]*Robert H. Smith School of Business, University of Maryland, College Park, MD, USA*

[2]*Simon Business School, University of Rochester, NY, USA*

[3]*Center for Digital Health and Artificial Intelligence (CDHAI), Carey Business School, Johns Hopkins University, MD, USA*

## Abstract

Electronic health records (EHRs) serve as an essential data source for the envisioned artificial intelligence (AI)-driven transformation in healthcare. However, clinician biases reflected in EHR notes can lead to AI models inheriting and amplifying these biases, perpetuating health disparities. This study investigates the impact of stigmatizing language (SL) in EHR notes on mortality prediction using a Transformer-based deep learning model and explainable AI (XAI) techniques. Our findings demonstrate that SL written by clinicians adversely affects AI performance, particularly so for black patients, highlighting SL as a source of racial disparity in AI model development. To explore an operationally efficient way to mitigate SL's impact, we investigate patterns in the generation of SL through a clinicians' collaborative network, identifying central clinicians as having a stronger impact on racial disparity in the AI model. We find that removing SL written by central clinicians is a more efficient bias reduction strategy than eliminating all SL in the entire corpus of data. This study provides actionable insights for responsible AI development and contributes to understanding clinician behavior and EHR note writing in healthcare.

## 1. Introduction

In recent years, the application of artificial intelligence (AI) in healthcare has experienced exponential growth. Healthcare is being reshaped both clinically (Rajpurkar et al. 2022), with AI whose performance is comparable to medical experts in detecting diseases (Esteva et al. 2017, Hannun et al. 2019), and operationally (Lou and Wu 2021), where the important yet largely unfulfilled opportunity of medical AI is attracting substantial investment. With 82% of healthcare-related executives in the industry expecting an aggressive adoption of AI (Landi 2021), the market value of healthcare AI is projected to reach $41.70 billion by 2027[1]. The benefits of adopting AI in healthcare are significant, estimated to reduce healthcare spending by 5% to 10%, or $200 billion to $360 billion a year[2].

Among AI models, Natural Language Processing (NLP) is especially valued as the next revolution in healthcare, with large language models such as ChatGPT (Chat Generative Pre-training Transformer) achieving remarkable performance and landmark achievements, including passing the medical licensing exam (Kung et al. 2023) and authoring scientific papers (Stokel-Walker 2023). In healthcare, 80% of the information is recorded in unstructured text such as clinical notes digitized in electronic health records (EHRs) (Negro-Calduch et al. 2021). These data, capturing the physician's expertise and diagnostic and treatment decisions, together with observations about the patient, represent a vast untapped resource that contains substantial economic values (Atasoy et al. 2018) and can now be leveraged through the application of NLP (Sarzynska-Wawer et al. 2021, Lee et al. 2020, Alsentzer et al. 2019). Models such as ClinicalBERT (Clinical Bidirectional Encoder Representations from Transformers) (Huang et al. 2019) and BEHRT (BERT for EHR) (Li et al. 2020) incorporate all available data tokens (or "words") to create medical predictions, without considering the unwanted non-medical patterns embedded in the data. However, EHR data, especially the free text in clinical notes, do not solely reflect objective clinical and medical facts; as has been shown recently, behavioral patterns and even their biases in clinical practices may be recorded in the training data (Ganju et al. 2020). These are key factors to be reckoned with in AI fairness (Schönberger 2019, Buolamwini and Gebru 2018, Bolukbasi et al. 2016).

---

[1] Source: https://www.prnewswire.com/news-releases/global-ai-in-healthcare-market-report-2022-surging-investment-in-ai-deals-in-healthcare-boosts-sector-301658255.html, accessed on Mar 19, 2023

[2] Source: https://www.healthcaredive.com/news/artificial-intelligence-healthcare-savings-harvard-mckinsey-report/641163/, accessed on Mar 19, 2023

Recent policy discussions have noted that with the growing impact of AI on business and society, it is even more critical to ensure that AI is fair and equitable and treats all "populations" in an equivalent manner (Schwartz et al. 2022, The White House 2022, Mehrabi et al. 2021). This is especially important for healthcare, as health inequity can have life-changing consequences. Throughout history, racial biases and disparities have been major concerns in healthcare and have the potential to be exacerbated by digital technologies and AI (Agarwal et al. 2022). Chen et al. (2021) summarize the racial inequalities in various aspects of clinical practice revealed by recent AI models. Tamayo-Sarver et al. (2003) document that black patients were less likely to be provided opioids than whites and Latinos. Dresser (1992) highlights the racial bias in medical research and notes that a majority of medical findings are based on conclusions drawn from white male samples that exclude minority patients. Even after years of effort, and although explicit racial discrimination is prohibited in healthcare practice, the implicit bias persists in the healthcare system (Baird et al. 2022, Barr 2019). The advancement of digitalization and AI, while doubtless bringing various benefits, is also amplifying the potential harm and impact of implicit biases by enshrining them in software (Fu et al. 2021). There is a rising concern that EHRs may propagate or even magnify implicit biases (Sun et al. 2022). These biases are likely embedded in EHR data used to train AI, which clearly affect AI performance.

In this study, we focus on a specific type of bias in EHR notes, stigmatizing language (SL), which reflects clinicians' implicit bias towards patients. An example of SL in EHR notes is shown in Figure 1, with pejorative words such as "abuser" and "noncompliant" in the text. Previous research has documented the widespread existence of SL in EHR notes, with different racial groups being affected disproportionately. For example, the EHR notes of black patients contain significantly more SL than the EHR notes of white patients (Himmelstein et al. 2022).

> "…pt is a long time **abuser** of etoh, s/p CVA in [**2184**] w/ minimal sequelae. pt has been **noncompliant** for years regarding medical care, etoh addiction, etc. pt is estranged from all family, and has a mentally disabled girlfriend as well (the witness of the pt's fall)…"

**Figure 1.** An example of an EHR note that contains SL

We conduct a series of experiments to examine how SL affects AI performance and fairness in a classic clinical prediction task, i.e., mortality prediction, for ICU patients. Specifically, we first examine how the presence of SL in the testing data alters mortality prediction outcomes of a Transformer-based model that has been trained on EHR notes. Second, we focus on AI fairness by investigating whether black patients are disadvantaged by the AI's predictions, and if SL in EHR notes is associated with such racial disadvantage. Third, we examine whether removing SL from the training data can be a potential solution to help reduce the influence of SL. More importantly, we take into consideration the social conditions underlying the data generation process, as reflected in the clinical teams responsible for patient care, and explore whether central clinicians write more SL, and if removing their SL can be a better strategy to improve AI performance and fairness.

We use data from a well-known and sizable EHR dataset, Medical Information Mart for Intensive Care III (MIMIC-III), to conduct our analysis. The MIMIC-III data consists of de-identified EHR records associated with over 60,000 intensive care unit (ICU) admissions. The data was collected from the Beth Israel Deaconess Medical Center between 2001 and 2012. It contains a wide range of attributes related to patients, such as free-text clinical notes, demographic attributes, the ID of the clinician who treated the patient, admission and discharge records, and death time (if applicable). We utilize the data in the following ways. First, we use the free-text clinical notes as the textual features to feed a Transformer-based model for mortality prediction, where the label is whether the patient was deceased in this admission, obtained from the death time attribute. Second, to examine the impact of SL, we utilize a list of SL keywords[3] identified by previous research to determine if a note contains SL (Himmelstein et al. 2022), as shown in Table 1. By varying the presence of SL in the data during testing and training the model, we are able to understand the impact of SL on AI performance.

---

[3] Due to the unique context of EHR notes, the presence of certain keywords that seem neutral in general contexts may be harmful in their clinical usage. For example, the term "user" can be considered SL when referring to a patient with a history of substance use, as it can perpetuate negative stereotypes and judgments.

**Table 1**. Keyword list of SL

| Keyword List of SL |
|---|
| 'adherence', 'nonadherent', 'compliance', 'unwilling', 'abuse', 'belligerent', 'drug seeking', 'abuser', 'difficult patient', 'refused', 'refuses', 'noncompliance', 'argumentative', 'cheat', 'abuses', 'malingering', 'user', 'secondary gain', 'in denial', 'refuse', 'compliant', 'substance abuse', 'nonadherence', 'degenerate', 'drug problem', 'combative', 'fake', 'been clean', 'noncompliant', 'addicted', 'narcotics', 'habit', 'adherent' |

Third, the recorded ethnicity of patients allows us to evaluate AI fairness. Specifically, if the model performs unequally on the black and white patient instances or subsets, we can conclude the existence of AI racial biases. Lastly, since every EHR record has the unique ID numbers of the clinicians who treated the patient, we can construct a clinician collaborative network, where the nodes are clinicians, and there is an edge if two clinicians have served the same patient. Whether a clinician has written SL in the data is incorporated as a binary node attribute. Such a network allows us to examine the association between clinicians' centrality in care teams and their SL writing behavior. If they are positively associated, we can consider removing central clinicians' SL as a better strategy to mitigate SL's impact.

Our analysis yields four novel findings. First, using the leave-one-out and input reduction strategies of explainable AI (XAI), we find that SL can hinder the performance of a trained AI model for mortality prediction. Second, we utilize adversarial perturbation approaches to demonstrate that the trained model exhibits racial biases, which are associated with the presence of SL. Specifically, telling the model that the patient is Caucasian has almost no effect on the prediction, while telling that the patient is black decreases the predicted probability by 15.97%. However, when SL is removed, the racial gap in the model's predicted probability almost disappears, suggesting that SL is associated with the model's racial bias. Global explanations obtained from global leave-one-out strategy and global adversarial perturbations on all testing examples yield consistent findings. Third, we explore the training set and find that removing SL helps improve the model performance and narrows the racial gap in mortality prediction. In particular, using the first 24 hours of ICU admission as an observation window, removing SL reduces the gap between the model's predicted F1 scores for black patients versus white patients from 2.97% to 0.05%. Fourth, our results from social network analysis provide evidence that central clinicians are more likely to write SL in EHR notes, and removing the SL in their notes can further improve the model performance in all scenarios, especially for black patients.

This study makes several contributions to the general literature on AI in healthcare and that specific to health equity and racial disparities. As the digitization of healthcare data advances, more and more AI models are being developed to utilize complex EHR data for clinical outcome prediction (Seinen et al. 2022). Compared with structured data, clinical notes are now increasingly used as source data due to the breakthroughs in NLP, such as large language models (Patel and Lam 2023). However, the potential racial disparity issues of AI models using clinical notes remain underexplored. In this study, we identified a striking phenomenon that calls for extra caution when developing and implementing AI models in clinical decisions that rely on clinical notes. Previous research has expressed concerns about the racial disparity of AI mainly because of the human biases embedded in the clinical text that can be inherited by AI (Posner and Fei-Fei 2020). This study sheds nuanced light on these concerns by showing that even if the training data is not bothered by systematic discrimination, the trained AI model could still create a significant racial disparity.

We further contribute by introducing a specific type of clinical text, SL, to the field of AI model development. Compared with other linguistic patterns, the use of SL in clinical notes is a unique presentation of implicit biases (Park et al. 2021a). Our findings show that although SL is generally examined in healthcare practices, it should also be considered in developing AI models with NLP components. To our knowledge, this is the first study that examines how SL affects AI performance, especially with regard to racial disparities. Finally, we connect social network literature with AI development. The recent development of AI models is heavily influenced by the new models in computer science, with fewer organizational and work environment elements being incorporated, which are especially critical in real business use than experimentation on data (Enholm et al. 2022). This study combines the social network literature that reflects the behavioral patterns of different medical professionals and identifies the impact of structural positions in the network (e.g., centrality). In line with the findings of network analyses, we design a novel approach to improve unbiasedness while retaining performance.

## 2. Literature Review and Hypothesis Development

### 2.1 Race in Medical Decision Making and AI

In the United States, the Census Bureau defines five distinct races: White, Black or African American, American Indian or Alaska Native, Asian, and Native Hawaiian or Other Pacific Islander.[4] This categorization emphasizes the original geographic region of one's ancestry (Risch et al. 2002), reflecting distinct genetic heritage. The variable "race" pervades medical decision making (MDM) in a variety of ways. Population genetic studies have uncovered race-related genetic variations, and researchers have constructed ancestral-tree diagrams supporting the grouping within races (Bowcock et al. 1991, Bowcock et al. 1994, Calafell et al. 1998). This is also verified at the genetic level as the delineation of genetic clusters is found to be associated with the racial categories (Mountain and Cavalli-Sforza 1997, Stephens et al. 2001, Wilson et al. 2001, Rosenberg et al. 2002). Since there are biological differences across races, racial information may legitimately be useful for clinical purposes and is often included in MDM. For example, a mutant allele, C282Y, that occurs with high frequency in northern Europeans but is absent in the non-white population, is medically associated with hemochromatosis (Merryweather-Clarke et al. 2000). Similar patterns have been found for other conditions, including venous thromboembolic disease (5 percent of white people and ⩽1 percent for East Asians and Africans) (Ridker et al. 1997, Shen et al. 1997), susceptibility to Crohn's disease (corresponding gene in whites but not found in Japanese) (Hugot et al. 2001, Yamazaki et al. 2002), and the famous heterozygous CCR5–delta32 variant, which protects against HIV (present in 25% of white people but not in other races) (Stephens et al. 1998), as well as the heterogeneity in responding to medicine and other medical treatments, such as the toxic effect of N-acetyltransferase 2 (slow-acetylator phenotype 14% of East Asians, 34% of blacks and 54% of whites) (Yu et al. 1994).

---

[4] Source:
https://www.census.gov/quickfacts/fact/note/US/RHI625221#:~:text=OMB%20requires%20five%20minimum%20categories,report%20more%20than%20one%20race. accessed on Mar 19, 2023

In light of this evidence, it is reasonable to conclude that race and ethnicity capture genomically unique features that are valuable for diagnostic and prognostic clinical practice (Burchard et al. 2003). Today, with the completion of the human genome project in 2003 and the declining cost of DNA sequencing, an individual's entire genome is available to physicians, resulting in a growing interest in precision medicine. Furthermore, the breadth of information utilized in MDM has expanded to include the social determinants of health, as racial information is highly correlated with various social and economic factors.

However, although their inclusion might be appropriate and warranted in certain situations, the clinical benefit of including race in MDM must be carefully evaluated against potential harm. The presence of race-based health inequities in society is widely recognized, and the inclusion of race may also result in unintended adverse consequences related to accentuating race-based disparity. For example, using race in MDM may reinforce negative stereotypes and increase the likelihood of stigma and discrimination against certain racial groups. If race is used to identify individuals at higher risk for substance abuse, this may lead to increased scrutiny and suspicion of individuals from those racial groups, even if they do not have a history of substance abuse.

With the growing application of AI in healthcare, the prospect of exacerbating disparities has become a mounting concern in society (Char et al. 2020, Paulus and Kent 2020, Chen et al. 2021). Recent studies call for a reconsideration of using racial information (Vyas et al. 2020), and the World Health Organization (2021) released a report highlighting the same concern. Data and algorithms are fundamentally artifacts created by humans, and therefore reflect individual, social, and institutional biases. If an AI is directly trained on the EHR data, it inherits the racial bias from human doctors and may recommend biased actions, such as referring black patients less to critical treatments that save their lives. Given that a wide variety of human biases are created by social and economic environments, the healthcare system, and individual health professionals, they are widely prevalent and deeply embedded in clinical data. In this case, incorporating racial information could make an AI model inevitably racially biased if directly trained on real clinical data.

In addition, AI's lack of capability for human causal reasoning makes it potentially more racially injurious. Machine learning models directly extract association patterns from a complex combination of various input features. Without sufficient guidance, healthcare AI models may misuse certain patterns. Such adverse outcomes are vividly documented by Obermeyer et al. (2019), who show that the racial bias in healthcare machine learning models results from using health costs as a proxy for health needs. Lower spending on healthcare by black people is not because of lower needs but rather socioeconomic factors that constrain the fulfillment of these needs. Relying on historical data, the AI falsely interprets lower spending as a signal of better health than equally sick white patients. Similarly, in the NLP field, Zhang et al. (2020) show that if the contextual language models are trained on scientific articles, they tend to recommend prisons for violent black patients and hospitals for violent white patients in completing clinical note templates.

## 2.2 Implicit Bias, Stigmatizing Language, and AI

MDM studies have revealed the existence of a multitude of implicit biases, including those associated with race, such as the prejudice that black people are violent (Holroyd et al. 2016). Racial biases are prevalent in the healthcare system (Chapman et al. 2013), even including Intensive Care Units (ICU) (FitzGerald and Hurst 2017, Martin et al. 2016), representing one of the most common forms of implicit bias. Such biases may be reflected in actions, such as being less attentive to a black patient or dismissing patient fears and preferences (Beck et al. 2022), or in language, such as writing discriminatory language against certain groups of patients (Park et al. 2021a). Such clinicians' biases can be recorded in EHR data, and AI models and biases intersect when data containing human bias are used for model development. Therefore, EHR data has become a core source of racial disparity in healthcare AI (Parikh et al. 2019).

To better understand the biases of medical providers and their impact on healthcare AI racial fairness, we focus on clinicians' freely expressed opinions captured in medical notes. From the perspective of AI development, because of the growing availability of EHR data and the breakthroughs of NLP AI, medical notes have become a critical input as training data for various AI models, such as models for predicting mortality, specific diseases, readmission to hospitals, and length of stay (Seinen et al. 2022). With the striking success of NLP models, such as the widely-acclaimed performance of ChatGPT, medical notes are no longer to be used for only extracting certain medical entities as subsequent model's input. Instead, complex deep learning models are being applied and evolving in making clinical predictions from medical notes, from traditional NLP models like TFIDF (term frequency-inverse document frequency), to basic artificial neural network models like RNN (recurrent neural networks) and CNN (convolutional neural networks), to the Transformer-based models in recent studies (Wen et al. 2020). As the lack of interpretability (black box issue) increases with the advancement of deep learning models, the impact of implicit racial biases in medical notes becomes more opaque and potentially more detrimental to racial fairness (Rudin 2019).

Within clinical notes, biases are expressed through what has been termed *stigmatizing language* (SL) (Himmelstein et al. 2022). SL is typically discrimination against an identifiable group of people, a place, or a nation (National Institute on Drug Abuse 2021). SL assigns negative labels, stereotypes, and judgments to certain groups of people, including inaccurate or unfounded thoughts such as they are dangerous, incapable of managing treatment, or at fault for their condition (Werder et al. 2022). The use of SL can aggravate clinicians' implicit biases and decrease patients' own sense of hope (Kelly et al. 2015). As contemporary AI empowered by deep learning is entering the field of medical notes-based medical prediction, it is becoming increasingly important to understand the impact of SL on state-of-the-art deep learning models. Our study addresses this by focusing on the impact of including and excluding SL in medical notes for a mortality prediction task in the context of an ICU using a Transformer based deep learning model.

In contrast to an AI developed for deciding whether to refer an end-stage renal disease patient to critical treatments (e.g., renal transplantation), the measure of the outcome for mortality prediction is theoretically likely to be less prone to the racial bias of the clinicians (McGarvey et al. 2007). While a patient referral for treatment could be biased due to the clinician's decision patterns, mortality is a more objective outcome than clinicians' subjective assessment, i.e., a physician's estimate of mortality risk. However, the use of SL signals the presence of bias, and similar to a clinician making sub-optimal treatment choices based on race, it is plausible that recommended paths of action for patients whose notes contain SL are clinically inferior. Thus, although our focus is on mortality that, on the surface, is an objective outcome, the endpoint of mortality is not immune to bias. Does the presence of SL improve or detract from the performance of an AI for mortality prediction? The answer to this question is not clear.

Echoing the heated debate in medicine regarding the use of racial information, the use of SL in state-of-the-art deep learning models, e.g., Transformer based models, could be potentially both beneficial and detrimental. On the one hand, clinicians may discriminate against certain racial groups during treatment, ultimately resulting in higher odds of mortality than other racial groups (Greenwood et al. 2020). In this case, if there is systematic discrimination against certain racial groups in the SL, SL is an effective predictor of the medical consequences (i.e., mortality) of biases in health practice. As a result, removing it from the deep learning model could cause information loss and reduce the performance of the mortality prediction. This is especially true for Transformer based deep learning models. Such models utilize a self-attention algorithm that learns global contexts of the entire set of inputs instead of only local contexts (e.g., surrounding words of a focal word), thereby being able to integrate more information (e.g., patterns in SL across different races) to make predictions (Vaswani et al. 2017). Therefore we test:

**Hypothesis 1a (H1a).** *Removing SL from medical notes can reduce the performance of AI models for mortality prediction.*

Alternatively, it is possible that although racial disparities are pervasive, they do not necessarily capture informative patterns in SL embedded in medical notes. In a comprehensive examination of SL in medical notes, Park et al. (2021a) suggest that SL could be about expressing clinicians' feelings (either positive or negative), which are not informative and may reflect biases. If SL is noninformative but without a clear systematic racial bias, it is fundamentally noise in the information. Prior research has discussed the negative impact of noise in medical notes (Miotto et al. 2016, Xiao et al. 2018 ), suggesting that including SL in an AI could impede its performance. In other words, removing noise in training data (i.e., the noninformative SL in medical notes) can effectively improve AI performance (Yang and Song 2010). Accordingly, we hypothesize that:

**Hypothesis 1b (H1b).** *Removing the SL from medical notes can improve the performance of AI models on mortality prediction.*

Among all racial groups, black patients are more vulnerable to racial disparities than white patients in clinical practice (Bailey et al. 2021). There are strong reasons to believe that black patients, compared with white patients, are also more disadvantaged in receiving high-performance predictions of a healthcare AI in our context of mortality prediction.

If SL is a reflection of racial discrimination in clinicians' healthcare practice, its effectiveness in improving mortality prediction depends on how informative it is as an indicator. As the information value of data is correlated with racial biases (Axt and Lai 2019), it is possible that the same word is interpretable as a reflection of bias toward black patients but as more personalized care for white patients. To illustrate, if a clinician makes an annotation of "non-compliant" for a white patient while specifically referencing a medication or appointment that the patient missed, it can be informative in predicting health outcomes for the patient. However, if the same word is utilized with black patients in a broad and stereotypical manner, it can reduce informativeness. In this case, including SL will better facilitate the prediction for white patients than black patients.

Even if SL is more of a noise than a predictor of the consequences of biases in health practice, the impact of the noise could also be heterogeneous and endanger racial unfairness in AI development. Since noise hurts the performance of the mortality prediction AI, the level and pattern of using SL in patients of different racial groups determine their level of disruption on AI's performance. Studies have documented that the medical records of black patients include more SL than those of white patients (Sun et al. 2022). Therefore, SL disrupts AI performance more for black patients than for white patients.

**Hypothesis 2 (H2).** *The existence of SL in medical notes is associated with racial disadvantage in AI fairness for black patients.*

## 2.3 Racial Disparity in AI and Clinician Social Structure

If as proposed in H2, SL yields racial disadvantages for certain populations and ultimately becomes a source of racial disparity in AI model development, mitigating the impact of SL is critical. To accomplish this, we focus on the underlying data generation process. Since SL is produced by individual clinicians, two natural questions are whether we can identify which clinicians are more likely to use SL, and how the elimination of SL produced by them could affect AI performance. We review the literature on the factors influencing implicit bias to guide our answer to these two questions.

Abundant research has suggested that implicit bias is associated with individual differences and environmental factors. The most common reason why people exhibit implicit bias is differences in individual attributes, such as individuals' attitudes, beliefs, and demographic characteristics (FitzGerald and Hurst 2017, Brownstein and Saul 2016). However, such individual attributes are usually personal traits that are difficult to influence. Prior research demonstrates that implicit biases derived from individuals' attributes are difficult to eliminate, even with conscious belief changes (Lai et al. 2016). In contrast to individual attributes, environmental factors are more amenable to intervention (FitzGerald et al. 2019). Therefore, focusing on the environmental factors that affect implicit bias can help us find actionable insights to curb the impact of SL.

Implicit biases are a mirror-like reflection of the environment within which the individuals are immersed (Dasgupta 2013). The environment here usually refers to the social structure formed by a group of people. Payne et al. (2017) suggest that implicit bias can emerge as individuals aggregate in crowds. With the greater intensity of interactions, as relational familiarity between individuals increases, they feel more comfortable using offensive language (Vingerhoets et al. 2013), and implicit biases are more reflected in their language. Clinicians are not immune to this pattern. Previous studies find that familiarity in the clinician-peer relationship is associated with the level of implicit bias (Centola et al. 2021). Therefore, it is plausible that clinicians are likely to increase their use of SL as their familiarity with peers rises. The relationship between familiarity with peers and SL use in notes is not likely to be dampened by the fact that clinicians interact with computers when writing medical notes. Previous research has found that individual guidelines governing offensive language may be loosened by increased familiarity, even in human-computer interaction scenarios (Park et al. 2021b). Furthermore, there is evidence that SL and implicit bias can transmit between clinicians via medical records (P Goddu et al. 2018). However, whether clinicians' SL writing is associated with their familiarity with peers remains understudied in the literature.

In our study, we examine if the social environment of individual clinicians, reflected in their familiarity with peers in the clinical team during the care-delivery process, is associated with SL, and how that affects AI performance. Specifically, we leverage social network analysis to study the impact of network centrality, as a measure of familiarity, on individual clinicians' SL writing behavior. Our focus on centrality is motivated by the essential nature of clinical work: the pervasive use of teams in healthcare delivery settings, and individual behavior in teams as a key element in effective care coordination (Rosen et al. 2018).

As noted earlier, there is a potential trade-off between the noise and information value of SL. If the information value of SL dominates, eliminating the SL produced by central clinicians could improve AI performance more than eliminating the SL produced by non-central clinicians. This is because, as clinicians become more familiar with their peers (i.e., centrality rises), they will exhibit more racial bias, which is negatively correlated with the information value of the SL. In this case, central clinicians' SL has lower information value than that of non-central clinicians, and removing it is more beneficial to AI performance. If SL is more of a noise, eliminating the SL produced by central clinicians is also a better strategy, because clinicians with higher centrality are expected to write more SL, which can hinder AI performance more. Therefore, we hypothesize:

**Hypothesis 3 (H3).** *Eliminating SL written by central clinicians improves AI performance and fairness in mortality prediction more than eliminating all SL.*

## 3. Research Design and Procedure

We investigate our research questions by developing a Transformer-based deep learning model using the MIMIC-III Clinical Database and applying XAI techniques to understand SL's effects. As illustrated in Figure 2, our research framework comprises three major components: Data and Preprocessing, Mortality Prediction, and Evaluation and Explanation. We explain each component in detail next.

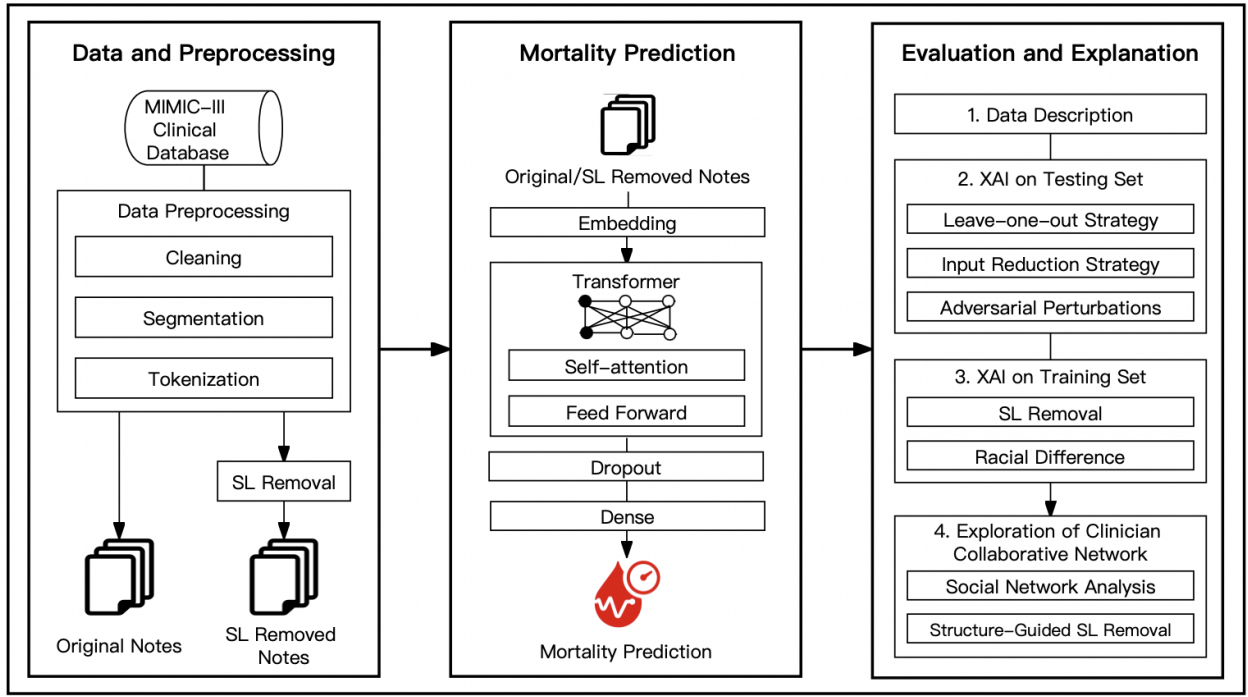

**Figure 2.** Our proposed research framework

## 3.1 Data and Preprocessing

Deep-learning-based AI models usually require sizable labeled data to perform. To this end, medical researchers have made large-size EHR datasets available to promote clinical predictive model development. If SL truly has negative impacts, various models relying on EHR notes are subject to biases. Among existing EHR datasets, MIMIC-III (Medical Information Mart for Intensive Care III) is the most widely used for model development in previous studies (Johnson et al. 2016). Therefore, we follow the literature to use MIMIC-III for a well-defined task, mortality prediction, by using either all records of ICU stays (i.e., full dataset) or the first 24 hours of ICU stays (i.e., 24-hour subset) (Seinen et al. 2022).

We employ the same preprocessing strategy as that in existing literature, such as the exclusion of discharge notes and token segmentation, to transform the complete data into a clean and meaningful input to AI models (Chen et al. 2019). Details of our data preprocessing steps are reported in Online Appendix B.

Identifying SL in EHR notes is a context-specific task requiring strong domain knowledge. Medical literature has developed and maintained a keyword list that is widely used to identify (and to further eliminate) the SL specifically in EHR notes (Himmelstein et al. 2022, Association of Diabetes Care and Education Specialists 2021, National Institute on Drug Abuse 2021). As a result, beyond the original data, we create another version by simply removing the SL keywords to examine the impact of SL on the subsequent predictive model development.

## 3.2 Mortality Prediction

To perform the mortality prediction task, we utilize a state-of-the-art deep learning model, Transformer, which is the core of the emerging GPT models (e.g., ChatGPT and GPT-4[5]). As the most advanced deep learning model, Transformer has been used by prior research and achieved state-of-the-art performance on clinical prediction tasks using EHR notes (Wen et al. 2020). A common practice in deep learning model development is transfer learning, where new models are developed on top of an existing pre-trained model. While previous models (e.g., ClinicalBERT) have achieved state-of-the-art performance on clinical prediction tasks, they were pre-trained on sizable EHR notes, where MIMIC-III is a popular source of pre-training data (Seinen et al. 2022). Therefore, our question of the potential racial bias of SL applies to the pre-trained models. In this case, to fully investigate our research question, it requires a clean development of Transformer models without transfer learning, as transfer learning on such pre-trained models for the focal mortality prediction on MIMIC-III data is subject to both information leakage and indirect impact of SL in pre-trained models. Therefore, we build a Transformer model from scratch without any pre-training. Using the EHR notes as input, the model makes a binary judgment on whether the patient is at risk of mortality, represented in a predicted probability. If the predicted probability surpasses 50%, a standard threshold, the patient is predicted as having a mortality risk. Our model is introduced in detail in Online Appendix B.

## 3.3 Evaluation and Explanation

---

[5] Due to ethical reasons, ChatGPT (both GPT-3.5 and GPT-4 versions) cannot predict specific clinical outcomes such as mortality. ChatGPT's standard response to such requests involves summarizing key elements in the EHR notes and recommending consultation with a licensed medical professional for a comprehensive assessment.

Arguably, identifying the effect of the model inputs on the decisions of AI is challenging due to the sophisticated decision-making process of AI models. Therefore, we apply XAI techniques to understand how SL affects AI performance during the testing and training phases. A comprehensive literature review on XAI can be found in Online Appendix A. Information Systems researchers have also proposed similar strategies, such as counterfactual explanations to understand AI decision-making (Fernandez et al. 2022, Martens and Provost 2014). The idea of counterfactual explanation is to consider the following situation: *If this input did not exist, what would have the AI model predicted?* The XAI techniques used in this study are largely consistent with this idea.

Our evaluation and explanation focus on three questions. First, if a Transformer based deep learning model is trained on EHR notes with SL, how will SL in its future prediction (the testing data) influence its racial disparity? Specifically, we use three XAI techniques to examine the impact of SL on the testing set: leave-one-out strategy (Li et al. 2016), input reduction strategy (Feng et al. 2018), and add sentence adversarial perturbations (Jia and Liang 2017). Second, is it beneficial to remove SL in the training phase? We then train the same model using EHR notes with SL and without SL. To quantify the impact of SL, we compare the model performance using four metrics: Accuracy, Precision, Recall, and F1 Score. Finally, we dive deeper to understand the behavioral root of the SL in EHR notes by conducting a formal social network analysis of clinicians. We specifically focus on clinicians with high centrality. Given their high degree of connections in the network, we aim to understand the patterns and impact of their SL usage on racial disparity in AI development.

## 4. Results

### 4.1 Data Description

MIMIC-III database is a widely used de-identified dataset focusing on critical care units. It includes more than 40,000 patients in the period 2001-2012 from the Beth Israel Deaconess Medical Center (Johnson et al. 2016). Despite the rich information, the raw MIMIC-III data is not directly usable since it is noisy and complex in structure. To this end, previous researchers have developed open-source pipelines to transform the raw MIMIC-III data into data structures that are suitable for clinical prediction tasks (Wang et al. 2020). Such pipelines also allow the direct comparison between different methods on the same clinical prediction tasks. In this study, we follow an open-source pipeline MIMIC-Extract (Wang et al. 2020) to transform the raw data from the MIMIC-III database, leading to our complete data, and we focus on the EHR notes in combination with the mortality results.

**Table 2**. Statistics of complete data

| Dataset | Patients | Notes | # of Notes Per Patient |
|---|---|---|---|
| **24 hours mortality** | 403 | 2,901 | 7.20 |
| **24-48 hours mortality** | 405 | 4,121 | 10.18 |
| **>48 hours mortality** | 2,335 | 75,560 | 32.36 |
| **Sum of mortality** | 3,143 | 79,681 | 25.35 |
| **Sum of survived** | 29,931 | 489,817 | 16.36 |
| **All** | 33,074 | 569,498 | 17.22 |

**Table 3**. Statistics of preprocessed data

| Dataset | # of SL Notes | # of Non-SL Notes | # of Notes |
|---|---|---|---|
| **Full** | 121,765 | 444,832 | 566,597 |
| **24 hours** | 11,661 | 123,862 | 135,523 |

| Dataset | Patients | # of Notes with Label = 1 | # of Notes with Label = 0 |
|---|---|---|---|
| **Full** | 32,671 | 76,202 | 490,395 |
| **24 hours** | 28,565 | 12,913 | 122,610 |

The statistics of the complete data are shown in Table 2. It comprises 569,498 EHR notes of 33,074 patients. Each patient has 17.22 notes on average and has only one ICU admission. The overall mortality rate is 9.5%, and 98.78% of patients survived the first 24 hours of their ICU stay, whereas 8.28% of patients survived 24 hours but subsequently deceased during their stay. The overall mortality rate conditional on surviving the first 48 hours is 2.44%. Table 3 illustrates the statistics of our preprocessed data. Compared with the complete data, the full dataset has 566,597 notes, with 2,902 notes removed during preprocessing. The full dataset covers 98.78% of patients, including all patients who survived the first 24 hours. The 24-hour subset only contains the EHR notes of the first 24 hours, and it covers 86.37% of all patients.

The number and percentage of EHR notes that contain SL in the data are presented in Table 4. On average, 21.49% of EHR notes contain SL, while black patients are 1.86% more likely to have SL in their EHR notes compared to white patients. This racial difference is consistent with prior studies (Himmelstein et al. 2022, Sun et al. 2022). The statistics suggest that there is a noticeable proportion of SL in the EHR notes, and SL is a vehicle of clinicians' bias toward patients.

**Table 4**. Number of EHR notes by SL and ethnicity

| Ethnicity | # of Stig Notes | Percentage | # of Non-stig Notes | Percentage |
|---|---|---|---|---|
| Black | 9,705 | 24.23% | 30,350 | 75.77% |
| White | 92,598 | 22.37% | 321,423 | 77.63% |
| Hispanic | 4,246 | 22.72% | 14,444 | 77.28% |
| Asian | 2,669 | 19.25% | 11,193 | 80.75% |
| Others | 12,547 | 15.69% | 67,422 | 84.31% |
| Sum | 121,765 | 21.49% | 444,832 | 78.51% |

## 4.2 Experimental Results: XAI on the Testing Set

### 4.2.1 XAI: Leave-one-out Strategy

**Table 5.** Results of the leave-one-out strategy

| Input | Probability | Improvement | Word importance | At Risk? |
|---|---|---|---|---|
| Transferred from outside hospital via [**Location (un) **]. Pt. apparently fell at home down [**5-4**] steps and had a +loc. When he aroused he was very ***combative***. | 48.23% | - | - | NO |
| Transferred from outside hospital via [**Location (un) **]. Pt. apparently fell at home down [**5-4**] steps and had a +loc. [MASK] he aroused he was very ***combative***. | 44.55% | -3.68% | #2 | NO |
| Transferred from outside hospital via [**Location (un) **]. Pt. apparently fell at home down [**5-4**] steps and had a +loc. When [MASK] aroused he was very ***combative***. | 46.34% | -1.89% | #4 | NO |
| Transferred from outside hospital via [**Location (un) **]. Pt. apparently fell at home down [**5-4**] steps and had a +loc. When he [MASK] he was very ***combative***. | 27.19% | -21.04% | #1 | NO |
| Transferred from outside hospital via [**Location (un) **]. Pt. apparently fell at home down [**5-4**] steps and had a +loc. When he aroused [MASK] was very ***combative***. | 46.16% | -2.07% | #3 | NO |
| Transferred from outside hospital via [**Location (un) **]. Pt. apparently fell at home down [**5-4**] steps and had a +loc. When he aroused he [MASK] very ***combative***. | 50.28% | 2.05% | #6 | YES |
| Transferred from outside hospital via [**Location (un) **]. Pt. apparently fell at home down [**5-4**] steps and had a +loc. When he aroused he was [MASK] ***combative***. | 49.55% | 1.32% | #5 | NO |
| Transferred from outside hospital via [**Location (un) **]. Pt. apparently fell at home down [**5-4**] steps and had a +loc. When he aroused he was very [MASK]. | 59.76% | 11.53% | #7 | YES |

We first apply the leave-one-out strategy to examine whether SL in the testing set affects our AI model performance. We start with an example to showcase the effect of SL. Specifically, we extract a positive example (i.e., the patient was eventually deceased during the ICU stay) from the test set to explore how SL affects mortality prediction results. As shown in Table 5, the model's mortality prediction for this patient is 48.23% using the original note, lower than the 50% threshold to make the correct prediction. This result suggests that the patient would not be at risk of mortality, which is an erroneous prediction. Notably, removing "combative," an SL word, leads to an 11.53% increase in the predicted probability, which results in a correct prediction. Removing any other word does not yield similar improvements. In particular, removing "aroused" reduces the mortality prediction by 21.04%, suggesting that this word is an important feature (ranked #1) for mortality prediction in this case. The results suggest that SL might impede the performance of an AI model trained on EHR notes in mortality prediction, and removing the SL could be an effective solution.

**Table 6.** Results of global leave-one-out strategy on all SL samples

| Condition | Precision | Recall | Compared with Original Recall | Recall for Positive Cases | Compared with Original Recall for Positive Cases | F1 |
|---|---|---|---|---|---|---|
| Original | 66.64% | 73.46% | - | 57.26% | - | 69.08% |
| SL Removal | 66.52% | 73.82% | 0.36% | 58.33% | 1.07% | 69.14% |
| Random Removal | 66.58% | 73.43% | -0.03% | 57.15% | -0.11% | 69.06% |

There may be concerns that the results only apply to the chosen example or that the findings are merely coincidental, and there is no difference in the effect of removing SL and random words. To address these concerns, we employ the leave-one-out strategy on our testing set globally and compare the average predictive probabilities under three conditions: a) utilizing the original notes, b) removing SL, and c) removing an equivalent number of random non-SL words. To ensure that the results of random removal are not coincidental, we perform random removal 100 times and report their average. In addition, for ease of comparison, we focus on samples containing SL. This is because we evaluate the same trained model in all conditions, and the predictions for samples without SL would not change with SL removal. Thus, by concentrating on samples containing SL, we can investigate the impact of SL on prediction results more efficiently.

As shown in Table 6, although removing SL from the testing set only has a marginal impact on the precision and F1 score, it does lead to an increase in the recall rate (0.36%), especially a noticeable increase in the recall rate for positive cases (1.07%). It is worth noting that an increase in the recall rate in medical predictions can help models capture more patients at risk, thereby reducing malpractice and moral hazard. In contrast, randomly removing the same number of non-SL words almost leads to no change to the model performance. This suggests that SL in medical notes does not exhibit informative patterns concerning this ICU outcome and instead introduces noise into AI models. Our findings remain consistent across these multiple tests, supporting Hypothesis 1b and refuting Hypothesis 1a.

### 4.2.2 XAI: Input Reduction Strategy

**Table 7.** Results of the input reduction strategy

| Input | Probability | Improvement | Change Direction |
|---|---|---|---|
| Transferred from outside hospital via [**Location (un) **]. Pt. apparently fell at home down [**5-4**] steps and had a +loc. When he aroused he was very combative. | 48.23% | - | - |
| Transferred from outside hospital via [**Location (un) **]. Pt. apparently fell at home down [**5-4**] steps and had a +loc. When he aroused he was very ~~combative~~. | 59.56% | 11.33% | ↑ |
| Transferred from outside hospital via [**Location (un) **]. Pt. apparently fell at home down [**5-4**] steps and had a +loc. When he aroused he ~~was~~ very. | 59.91% | 11.68% | ↑ |
| Transferred from outside hospital via [**Location (un) **]. Pt. apparently fell at home down [**5-4**] steps and had a +loc. When he aroused he ~~very~~. | 59.02% | 10.79% | ↓ |
| Transferred from outside hospital via [**Location (un) **]. Pt. apparently fell at home down [**5-4**] steps and had a +loc. When ~~he~~ aroused he. | 53.78% | 5.55% | ↓ |
| Transferred from outside hospital via [**Location (un) **]. Pt. apparently fell at home down [**5-4**] steps and had a +loc. When aroused ~~he~~. | 49.30% | 1.07% | ↓ |
| Transferred from outside hospital via [**Location (un) **]. Pt. apparently fell at home down [**5-4**] steps and had a +loc. ~~When~~ aroused. | 43.90% | -4.33% | ↓ |
| Transferred from outside hospital via [**Location (un) **]. Pt. apparently fell at home down [**5-4**] steps and had a +loc. ~~aroused~~. | 30.62% | -17.61% | ↓ |

To further examine the robustness of the leave-one-out strategy results, we leverage the input reduction strategy to drop the least important word sequentially and observe the changes in the mortality prediction results. Table 7 illustrates the results using the same example. Consistent with our previous findings, removing "combative" can help increase the predicted probability, suggesting that SL is responsible for the erroneous prediction. In contrast, removing the rest of the sentence can reduce the predicted probability (except for "was," which only leads to a marginal increase of 0.35% in predicted probability). In particular, removing "aroused" leads to a 17.61% decrease in the predicted probability, underscoring its importance. These results further support our findings that SL could negatively affect AI models on clinical prediction, leading to inferior performance.

### 4.2.3 XAI: Adversarial Perturbations

**Table 8.** Results of add sentence adversarial perturbations

| Example | Input | Probability | Probability Change |
|---------|-------|-------------|--------------------|
| Original | Transferred from outside hospital via [**Location (un) **]. Pt. apparently fell at home down [**5-4**] steps and had a +loc. When he aroused he was very combative. | 48.23% | - |
| White + SL | Transferred from outside hospital via [**Location (un) **]. Pt. apparently fell at home down [**5-4**] steps and had a +loc. When he aroused he was very combative. ***Pt is a Caucasian***. | 48.18% | -0.05% |
| Black + SL | Transferred from outside hospital via [**Location (un) **]. Pt. apparently fell at home down [**5-4**] steps and had a +loc. When he aroused he was very combative. ***Pt is an African American***. | 32.36% | -15.87% |
| Non-SL | Transferred from outside hospital via [**Location (un) **]. Pt. apparently fell at home down [**5-4**] steps and had a +loc. When he was very [MASK]. | 59.76% | 11.53% |
| White + Non-SL | Transferred from outside hospital via [**Location (un) **]. Pt. apparently fell at home down [**5-4**] steps and had a +loc. When he was very [MASK]. ***Pt is a Caucasian***. | 61.07% | 12.84% |
| Black + Non-SL | Transferred from outside hospital via [**Location (un) **]. Pt. apparently fell at home down [**5-4**] steps and had a +loc. When he aroused he was very [MASK]. ***Pt is an African American***. | 63.62% | 15.39% |

Next, we employ add sentence adversarial perturbations, by inserting an adversarial sentence into the original input and analyzing how the model responds, to gain insights into the model's racial disparity caused by the use of SL. As shown in Table 8, when the "Caucasian" sentence (i.e., "Pt is a Caucasian.") is added, the model's prediction barely changes (-0.05%). However, a substantial drop can be observed when the "African American" sentence is added (-15.87%). Interestingly, this racial gap can be significantly narrowed if the SL is removed, with the addition of the "Caucasian" sentence and the addition of the "African American" sentence leading to a 12.84% increase and a 15.39% increase in the predicted mortality, respectively. Note that the patient in this example was deceased, suggesting that an increase in mortality prediction is beneficial. It is striking that, in this example, it is not just the racial gap being closed by simply removing SL words, but also the "African American" sentence is even higher than adding the "Caucasian" sentence. This provides initial evidence to support H2.

**Table 9.** Results of global add sentence adversarial perturbations

| Condition | Added Sentence | Precision | Recall | Compared with Original Recall | Recall for Positive Cases | Compared with Original Recall for Positive Cases | F1 |
|---|---|---|---|---|---|---|---|
| Original | None | 63.44% | 66.67% | - | 46.31% | - | 64.68% |
| White | *Pt is a Caucasian.* | 63.47% | 66.57% | -0.10% | 45.97% | -0.34% | 64.68% |
| Black | *Pt is an African American.* | 64.14% | 65.09% | -1.58% | 40.63% | -5.68% | 64.59% |
| SL | *Pt has a history of drug abuse.* | 63.64% | 65.57% | -1.10% | 42.68% | -3.63% | 64.47% |
| White + SL | *Pt is a Caucasian and has a history of drug abuse.* | 63.48% | 65.04% | -1.63% | 41.35% | -4.96% | 64.17% |
| Black + SL | *Pt is an African American and has a history of drug abuse.* | 64.09% | 63.65% | -3.02% | 36.55% | -9.76% | 63.86% |

One concern of add sentence adversarial perturbations is that the interpretations regarding SL are local and only valid for this selected example and our set of added sentences. To understand the results of adversarial perturbations globally, we add the adversarial sentence to all samples of the testing set to systematically examine the impact of SL on racial disparity in model performance. As illustrated in Table 9, the model changes little when adding the "Caucasian" sentence (0% in F1, -0.10% in recall rate, and -0.34% in recall rate for positive cases). Adding the "African American" sentence results in a noticeable drop in F1 (-0.09%), as well as a 1.58% and 5.68% drop in the recall rate and the recall rate for positive cases, respectively. Furthermore, we add an additional SL phrase[6], which indicates the patient has a history of drug abuse, to investigate how SL interacts with racial information. Adding additional SL further reduces the recall rate, while black patients are worse off, with a 4.8% greater drop in positive case recall than Caucasians (9.76% compared with 4.96%). The results are largely consistent with our previous findings. Echoing our discussion on the recall rate, a drop in the recall rate is dangerous for medical predictions, since models can create malpractice and moral hazard when patients at high mortality risk are missed.

Taken together, our results suggest that SL is detrimental to an AI model for mortality prediction, not only on the performance but also on the fairness. Therefore, H1b and H2 are supported on the testing set. While the two sets of results show the harm of SL in both model performance and racial fairness of AI, a natural solution is to remove the SL from the training set so that the model can get rid of the biases. Next, we investigate the effect of removing SL from the training data.

---

[6] The added phrase is "*and has a history of drug abuse*" which attempts to express that the individual has struggled with substance use in the past. However, to avoid perpetuating stigma, it is important to use more neutral and person-centered language such as "*and has a history of substance use disorder.*"

## 4.3 Experimental Results: XAI on the Training Set

**Table 10**. Results of removing SL from the training set

| Set ID | Dataset | Accuracy | Precision | Recall | F1-Score |
|---|---|---|---|---|---|
| 1 | Full (Original) | 81.57% | 63.44% | 66.67% | 64.68% |
| 2 | Full (SL Removed) | 86.37% | 69.56% | 64.44% | 66.36% |
| | **Improvement from SL Removal** | **4.80%** | **6.12%** | **-2.23%** | **1.68%** |
| 3 | Full - White patients (Original) | 81.85% | 62.77% | 66.30% | 64.08% |
| 4 | Full - Black patients (Original) | 82.90% | 61.56% | 67.80% | 63.39% |
| | **Racial Difference (Black-White)** | **1.05%** | **-1.21%** | **1.50%** | **-0.69%** |
| 5 | Full - White patients (SL Removed) | 86.75% | 68.79% | 64.12% | 65.91% |
| 6 | Full - Black patients (SL Removed) | 88.31% | 67.12% | 65.34% | 66.15% |
| | **Racial Difference (Black-White)** | **1.56%** | **-1.67%** | **1.22%** | **0.24%** |
| 7 | 24 Hours (Original) | 89.54% | 67.67% | 62.13% | 64.16% |
| 8 | 24 Hours (SL Removed) | 90.53% | 71.56% | 61.16% | 64.13% |
| | **Improvement from SL Removal** | **0.99%** | **3.89%** | **-0.97%** | **-0.03%** |
| 9 | 24 Hours - White patients (Original) | 89.83% | 67.14% | 62.25% | 64.11% |
| 10 | 24 Hours - Black patients (Original) | 92.17% | 61.56% | 60.76% | 61.14% |
| | **Racial Difference (Black-White)** | **2.34%** | **-5.58%** | **-1.49%** | **-2.97%** |
| 11 | 24 Hours - White patients (SL Removed) | 90.80% | 70.69% | 60.92% | 63.78% |
| 12 | 24 Hours - Black patients (SL Removed) | 93.67% | 67.31% | 61.56% | 63.73% |
| | **Racial Difference (Black-White)** | **2.87%** | **-3.38%** | **0.64%** | **-0.05%** |

To examine our hypotheses on the training set, we trained a Transformer model on 12 different settings and compared the relative performance, as shown in Table 10. On the full dataset, removing SL leads to a 1.68% increase in the F1 score (Set 1 and 2). We find that the Transformer model performs better on white patients than on black patients using the original notes (for both full data and 24 Hours data). In particular, there is a salient racial gap on the 24-hour subset (Set 9 and 10), where the model achieves a 2.97% higher F1 score for white patients than black patients. However, when SL is removed in Set 11 and 12, the racial gap almost vanishes (0.05%).

These results provide evidence that racial disparities in EHR notes can propagate to AI models, and SL is a core element associated with racial disparities. Removing SL could significantly improve AI performance and fairness. Therefore, our H1b and H2 are supported on the training set as well. These findings are critical to AI practitioners in clinical settings as SL is pervasively available in public and private EHR notes. While less attention has been paid to SL in previous model development, existing development and publicly accessible pre-trained models may preserve racial disparities caused by SL. Immediate actions are needed to prevent the racial disparities from being propagated via the hard-to-interpret deep learning models. In another thread, more understanding of the behavioral patterns of SL usage is called for to enable the design of strategies to address SL in EHR notes.

## 4.4 Experimental Results: Exploration of Clinician Collaborative Network

### 4.4.1 Social Network Analysis

To further explore a better strategy to address SL in EHR notes, we go beyond the simple removal of SL and focus on clinicians' structural positions in their collaborative network. Specifically, we apply social network analysis to investigate if the degree centrality of clinicians in their collaborative network affects their use of SL and the subsequent impact on AI performance. Our focus on centrality is motivated by prior research in social psychology, which suggests that social familiarity plays a significant role in shaping individual behavior and communication patterns (Vingerhoets et al. 2013, Sparrowe et al. 2001). In the in-patient healthcare context, as discussed in Section 2.3, teamwork is an integral part of the care-delivery process, with clinicians frequently collaborating and communicating with one another to provide optimal patient care. However, previous research suggests that while teamwork increases familiarity among clinicians, the familiarity can also make them feel more comfortable exhibiting implicit biases and using offensive language (Centola et al. 2021, Vingerhoets et al. 2013). Therefore, we expect that the extent to which a focal clinician is familiar with others in the team is positively associated with their SL usage in EHR notes.

Although we cannot directly observe individual clinicians' behavior due to data limitations, we can infer their behavior through their structural position in the clinician collaborative network. If we observe a positive relationship between centrality and SL, as predicted by previous theories and as we proposed, we can infer that central clinicians exhibit more SL writing behavior, thereby informing SL removal strategies. To this end, we constructed a collaborative network of clinicians, with the weight of an edge between two clinicians representing the number of patients they served together. Our data contain 1,492 clinicians and 215,168 edges. More details about the collaborative network can be found in Online Appendix C.

We first associate SL use with the degree centrality of clinicians. In a weighted network, the degree centrality is generalized and measured by node strength, the sum of the weights of all the edges a node has (Opsahl et al. 2010, Barrat et al. 2004). We define central clinicians as those whose node strength is above average and the rest as non-central ones. Figure 3 visually displays the collaborative networks in our data: white nodes are clinicians with SL, while the red nodes are those with no SL in their notes. As is easily discernible, clinicians who are more connected with other clinicians and have higher centrality are more likely to write SL in EHR notes. Specifically, among 1,357 clinicians who used SL in their notes, 74.43% of them are in the central clusters (i.e., 1,010 clinicians). In contrast, only 8.89% of the clinicians who never wrote SL are in the central clusters (i.e., 12 out of 135). In addition, we broke down the clinician category into physicians and nurses and found consistent patterns, as shown in Figure 4. Table 11 provides a more direct comparison between the SL and EHR notes written by central clinicians and others. Notably, central physicians wrote 13.87% more SL than non-central physicians.

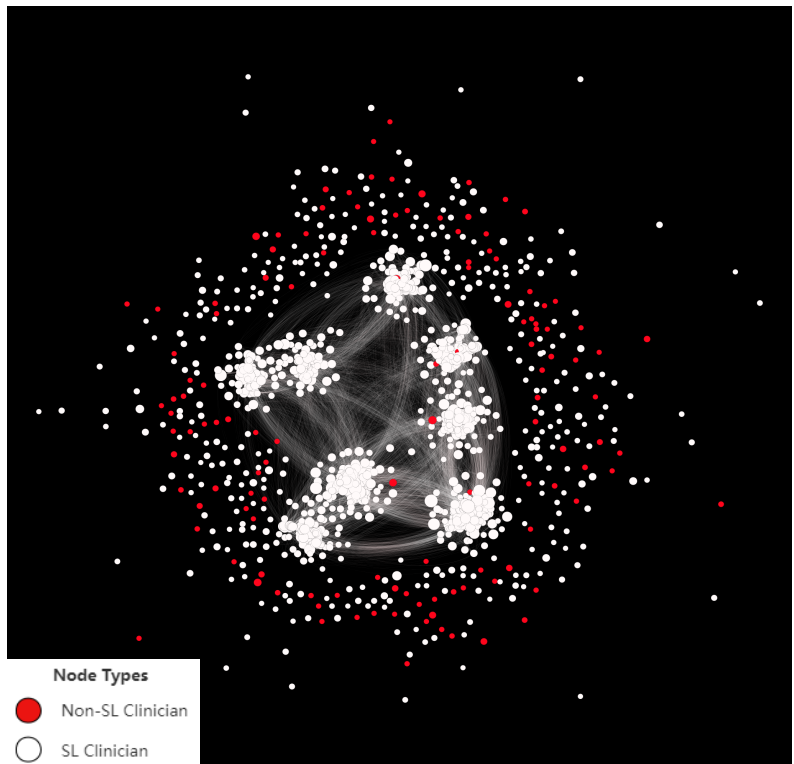

**Figure 3**. Social network analysis of all clinicians. **Note:** The OpenOrd layout is applied since it aims to distinguish clusters of large and undirected weighted graphs (Martin et al. 2011). To avoid

dense and overlapped edges in the visualization, we only show edges with weights in the top 10% (i.e., above 80), leading to 21,605 edges in this graph.

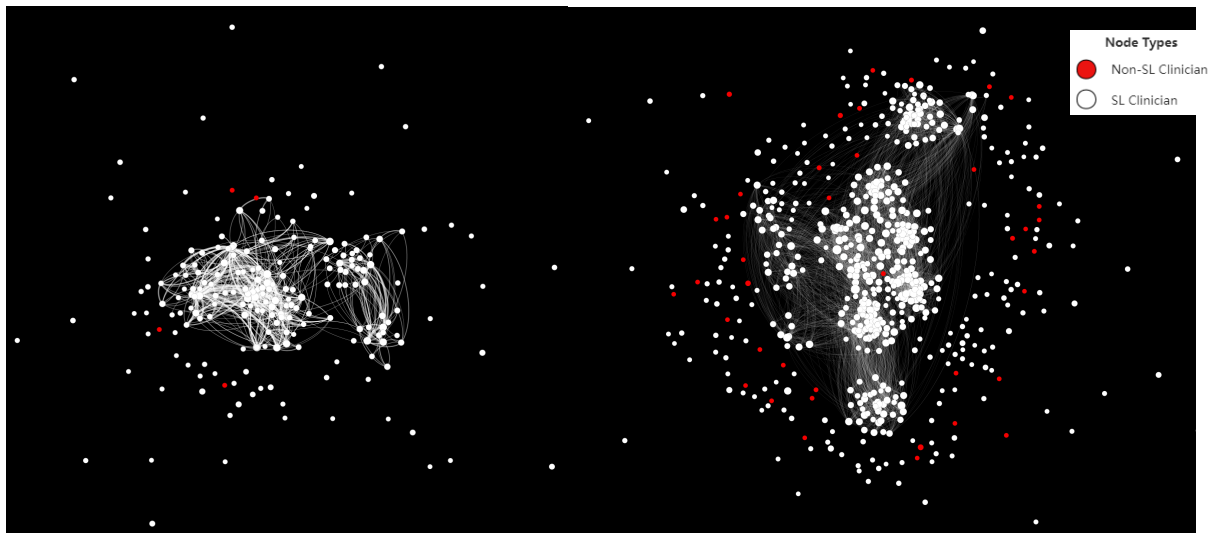

**Figure 4.** Social network analysis of physicians (left) and nurses (right)

**Table 11**. Statistics of SL in EHR notes by clinician centrality

| Measure\Centrality | Central | Non-central | Difference | Total |
|---|---|---|---|---|
| # of Clinicians | 1,022 | 470 | 552 | 1,492 |
| %SL Clinicians | 98.63% | 89.36% | 9.27% | 552 |
| # of Notes (proportion) | 348,781 (61.24%) | 111,140 (19.52%) | 237,641 | 569,498 (100%) |
| %SL Notes (scaled) | 20.32% | 6.45% | 13.87% | 21.49% |
| %SL of Physician Notes (scaled) | 15.97% | 4.45% | 11.52% | 25.98% |
| %SL of Nurse Notes (scaled) | 4.39% | 1.35% | 3.04% | 7.16% |

### 4.4.2 Structure-Guided SL Removal

Our SNA results suggest that the central clinicians write significantly more SL. One possible explanation is that central clinicians wield more influence due to greater access to resources and opportunities. Consequently, they may feel more at ease expressing themselves in EHR notes, resulting in increased use of SL in their notes. Alternatively, central clinicians are connected to more colleagues, thus, are exposed to more of their SL. In either scenario, central clinicians are more likely than non-central clinicians to write SL. However, such SL usage is not focused on conveying critical patient information related to mortality but reflective of their peer relationships and social environment, and therefore, it contains no systematic patterns that can help mortality prediction. Our results further support H1b, which shows that SL is disruptive to the model performance as there is no systematic pattern in SL. As such, instead of solely examining the quantity of SL used by central versus non-central clinicians, we should also consider the disruptive nature of SL used by central clinicians.

Correspondingly, with the SL used by central clinicians less predictable and more disruptive, it stands to reason that their SL is more responsible for the negative effect on AI's performance and racial fairness. In this case, removing SL written by central clinicians could further improve AI performance and fairness. The results are presented in Table 12. Interestingly, removing SL written by central clinicians achieves higher performance than removing all SL blindly, as indicated by the F1 score (+0.45% for the full sample and +0.37% for 24 Hours sample), suggesting that the SL used by central clinicians is disruptive to the AI's performance. This not only echoes our discussion that central clinicians' SL has non-systematic patterns but also implies that the SL used by non-central clinicians could potentially exhibit greater regularity.

In the comparison between different racial groups, black patients are significantly better off from the removal of only central clinician-authored SL. The F1 score increased by 2.12% and 4.98% on the full set and 24-hour subset, respectively. These findings support H3 and further unravel the behavioral root of SL's negative impact on both model performance and AI racial fairness. Our results provide guidelines for AI practitioners to better deal with SL in EHR notes when developing machine learning models. In addition, they also pave a way to discover new issues in training for AI development by searching the behavioral patterns of the data generation process.

**Table 12**. Results of removing SL of central clinicians

| Dataset | Accuracy | Precision | Recall | F1-Score |
|---|---|---|---|---|
| Full (SL Removed) | 86.37% | 69.56% | 64.44% | 66.36% |
| Full (Central SL Removed) | 88.09% | 75.73% | 63.49% | 66.81% |
| Full - White patients (SL Removed) | 86.75% | 68.79% | 64.12% | 65.91% |
| Full - White patients (Central SL Removed) | 88.49% | 74.84% | 63.11% | 66.35% |
| Full - Black patients (SL Removed) | 88.31% | 67.12% | 65.34% | 66.15% |
| Full - Black patients (Central SL Removed) | 90.86% | 75.89% | 64.74% | 68.27% |
| 24 Hours (SL Removed) | 90.53% | 71.56% | 61.16% | 64.13% |
| 24 Hours (Central SL Removed) | 89.42% | 67.41% | 62.68% | 64.50% |
| 24 Hours - White patients (SL Removed) | 90.80% | 70.69% | 60.92% | 63.78% |
| 24 Hours - White patients (Central SL Removed) | 89.61% | 66.85% | 62.28% | 64.04% |
| 24 Hours - Black patients (SL Removed) | 93.67% | 67.31% | 61.56% | 63.73% |
| 24 Hours - Black patients (Central SL Removed) | 93.72% | 66.98% | 70.96% | 68.71% |

## 5. Conclusion

In this study, we focused on the impact of bias reflected on stigmatizing language in EHR notes on AI performance and fairness. Our results suggest that SL can hinder the performance of an AI model for mortality prediction, mainly because of the noisy patterns in SL use. We found that SL yields racial disadvantages for black patients in AI model development and becomes a source of racial disparities, jeopardizing AI fairness. These findings are important for both training an AI model using SL EHR notes and applying a trained model to SL EHR notes. First, as the racial disparity associated with SL can propagate to AI models, removing SL from the training data can mitigate the impact of SL, improving AI performance and fairness. Second, even if an AI was trained on SL EHR notes, excluding SL in testing data can still reduce the racial gap. Finally, we situate the generation of SL in clinicians' behavioral patterns by performing social network analysis. SL from clinicians with high centrality in the network is responsible for the negative impact on both AI performance and AI fairness. We found that removing SL written by central clinicians is a better strategy to achieve racial equity and AI fairness than simply removing all SL.

The recent progress in large language models, such as ChatGPT, is a major engineering achievement in applying state-of-the-art models using high-quality data (Zhou et al. 2023). The core asset of ChatGPT is clearly the training dataset (Perrigo 2023), with the source of the data rapidly becoming the subject of a heated debate between OpenAI and Google (Hollister 2023). Given the lack of new algorithmic breakthroughs in the deep learning models for processing natural language since Transformers in 2017 (Vaswani et al. 2017), the quantity and quality of the training data have become especially critical for deep learning model development. Such a shift in focus on data makes healthcare AI development no exception. However, there are various healthcare-specific factors to be considered in developing clinical AI models. This is especially true for medical texts such as EHR notes, where factors influencing AI performance and fairness remain unstudied.

Our study fills the gap by identifying a striking phenomenon that the racial disparities in clinicians' SL in EHR notes can propagate to AI models, resulting in both low performance and racial unfairness in AI, thereby offering robust opportunities for future work. Our study makes contributions to both medical AI development and AI fairness in general. Information Systems researchers have been developing machine learning models utilizing EHR or related clinical data (Samtani et al. 2023, Yu et al. 2021, Bardhan et al. 2020, Zhang and Ram 2020). However, the negative impact of SL is absent in most clinical predictive model development. AI fairness is a popular research topic in general (Chen et al. 2021, Schönberger 2019, Buolamwini and Gebru 2018, Bolukbasi et al. 2016), and it is rising in business research (Kallus et al. 2022, Yue et al. 2022, Bjarnadóttir and Anderson 2020, Malik and Singh 2019). We adopt a distinctive perspective to examine how the biases in human behavior (SL) can negatively impact a broad scope of patients through its propagation in AI model development. Our work also contributes to the literature on SL in EHR (Himmelstein et al. 2022) by rooting the SL use in social network analytics. Extending the analyses of SL usage from the individual level to the organizational structural level provides a new avenue for understanding SL in clinical texts. Finally, research on human biases in clinical practice has a long history in Information Systems (Ganju et al. 2022, Lin et al. 2019). This study connects this stream of literature to the emerging topic of AI development. Practically, a meaningful mitigation strategy for reducing the harms of SL is provided. Removing SL written by central clinicians is shown to be effective in curbing the propagation of racial disparities in AI prediction, lending actionable insights to policymakers and industry practitioners for responsible AI development.

## References

Agarwal, R., Bjarnadottir, M., Rhue, L., Dugas, M., Crowley, K., Clark, J., & Gao, G. (2022). Addressing algorithmic bias and the perpetuation of health inequities: An AI bias aware framework. *Health Policy and Technology*, 100702.

Alsentzer, E., Murphy, J., Boag, W., Weng, W. H., Jindi, D., Naumann, T., & McDermott, M. (2019, June). Publicly Available Clinical BERT Embeddings. In *Proceedings of the 2nd Clinical Natural Language Processing Workshop* (pp. 72-78).

Association of Diabetes Care and Education Specialists. (2021). Language Guidance for Diabetes-Related Research, Education and Publications. Retrieved Apr 2, 2023,

https://www.diabeteseducator.org/docs/default-source/practice/educator-tools/HCP-diabetes-language-guidance.pdf?sfvrsn=22.

Atasoy, H., Chen, P. Y., & Ganju, K. (2018). The spillover effects of health IT investments on regional healthcare costs. *Management Science*, *64*(6), 2515-2534.

Axt, J. R., & Lai, C. K. (2019). Reducing discrimination: A bias versus noise perspective. *Journal of Personality and Social Psychology*, *117*(1), 26.

Bailey, Z. D., Feldman, J. M., & Bassett, M. T. (2021). How structural racism works—racist policies as a root cause of US racial health inequities. *New England Journal of Medicine*, *384*(8), 768-773.

Baird, A., Cheng, Y., & Xia, Y. (2022). Use of machine learning to examine disparities in completion of substance use disorder treatment. *PloS one*, *17*(9), e0275054.

Bardhan, I., Chen, H., & Karahanna, E. (2020). Connecting systems, data, and people: A multidisciplinary research roadmap for chronic disease management. *MIS Quarterly*, *44*(1), 185-200.

Barr, D. A. (2019). *Health disparities in the United States: Social class, race, ethnicity, and the social determinants of health*. JHU Press.

Barrat, A., Barthelemy, M., Pastor-Satorras, R., & Vespignani, A. (2004). The architecture of complex weighted networks. *Proceedings of the national academy of sciences*, *101*(11), 3747-3752.

Beck, A. S., Svirsky, L., & Howard, D. (2022). 'First Do No Harm': physician discretion, racial disparities and opioid treatment agreements. *Journal of Medical Ethics*, *48*(10), 753-758.

Bjarnadóttir, M. V., & Anderson, D. (2020). Machine learning in healthcare: Fairness, issues, and challenges. In *Pushing the Boundaries: Frontiers in Impactful OR/OM Research* (pp. 64-83). INFORMS.

Bolukbasi, T., Chang, K. W., Zou, J. Y., Saligrama, V., & Kalai, A. T. (2016). Man is to computer programmer as woman is to homemaker? debiasing word embeddings. *Advances in neural information processing systems*, *29*.

Bowcock, A. M., Kidd, J. R., Mountain, J. L., Hebert, J. M., Carotenuto, L., Kidd, K. K., & Cavalli-Sforza, L. L. (1991). Drift, admixture, and selection in human evolution: a study with DNA polymorphisms. *Proceedings of the National Academy of Sciences*, *88*(3), 839-843.

Bowcock, A. M., Ruiz-Linares, A., Tomfohrde, J., Minch, E., Kidd, J. R., & Cavalli-Sforza, L. L. (1994). High resolution of human evolutionary trees with polymorphic microsatellites. *Nature*, *368*(6470), 455-457.

Brownstein, M., & Saul, J. (Eds.). (2016). *Implicit bias and philosophy, volume 2: Moral responsibility, structural injustice, and ethics*. Oxford University Press.

Buolamwini, J., & Gebru, T. (2018, January). Gender shades: Intersectional accuracy disparities in commercial gender classification. In *Conference on fairness, accountability and transparency* (pp. 77-91). PMLR.

Burchard, E. G., Ziv, E., Coyle, N., Gomez, S. L., Tang, H., Karter, A. J., ... & Risch, N. (2003). The importance of race and ethnic background in biomedical research and clinical practice. *New England Journal of Medicine*, *348*(12), 1170-1175.

Calafell, F., Shuster, A., Speed, W. C., Kidd, J. R., & Kidd, K. K. (1998). Short tandem repeat polymorphism evolution in humans. *European Journal of Human Genetics*, *6*(1).

Centola, D., Guilbeault, D., Sarkar, U., Khoong, E., & Zhang, J. (2021). The reduction of race and gender bias in clinical treatment recommendations using clinician peer networks in an experimental setting. *Nature communications*, *12*(1), 6585.

Chapman, E. N., Kaatz, A., & Carnes, M. (2013). Physicians and implicit bias: how doctors may unwittingly perpetuate health care disparities. *Journal of general internal medicine*, *28*(11), 1504-1510.

Char, D. S., Abràmoff, M. D., & Feudtner, C. (2020). Identifying ethical considerations for machine learning healthcare applications. *The American Journal of Bioethics*, *20*(11), 7-17.

Chen, I. Y., Pierson, E., Rose, S., Joshi, S., Ferryman, K., & Ghassemi, M. (2021). Ethical machine learning in healthcare. *Annual review of biomedical data science*, *4*, 123-144.

Chen, I. Y., Szolovits, P., & Ghassemi, M. (2019). Can AI help reduce disparities in general medical and mental health care?. *AMA journal of ethics*, *21*(2), 167-179.

Dasgupta, N. (2013). Implicit attitudes and beliefs adapt to situations: A decade of research on the malleability of implicit prejudice, stereotypes, and the self-concept. *Advances in experimental social psychology*, *47*, 233-279.

Dresser, R. (1992). Wanted single, white male for medical research. *The Hastings Center Report*, *22*(1), 24-29.

Esteva, A., Kuprel, B., Novoa, R. A., Ko, J., Swetter, S. M., Blau, H. M., & Thrun, S. (2017). Dermatologist-level classification of skin cancer with deep neural networks. *nature*, *542*(7639), 115-118.

Feng, S., Wallace, E., Grissom II, A., Iyyer, M., Rodriguez, P., & Boyd-Graber, J. (2018). Pathologies of Neural Models Make Interpretations Difficult. In *Proceedings of the 2018 Conference on Empirical Methods in Natural Language Processing* (pp. 3719-3728).

Fernandez, C., Provost, F., & Han, X. (2022). Explaining data-driven decisions made by AI systems: the counterfactual approach. *MIS Quarterly*, *46*(3).

FitzGerald, C., Martin, A., Berner, D., & Hurst, S. (2019). Interventions designed to reduce implicit prejudices and implicit stereotypes in real world contexts: a systematic review. *BMC psychology*, 7(1), 1-12.

FitzGerald, C., & Hurst, S. (2017). Implicit bias in healthcare professionals: a systematic review. *BMC medical ethics*, *18*(1), 1-18.

Fu, R., Huang, Y., & Singh, P. V. (2021). Crowds, lending, machine, and bias. *Information Systems Research*, *32*(1), 72-92.

Ganju, K. K., Atasoy, H., & Pavlou, P. A. (2022). Do electronic health record systems increase medicare reimbursements? The moderating effect of the recovery audit program. *Management Science*, *68*(4), 2889-2913.

Ganju, K. K., Atasoy, H., McCullough, J., & Greenwood, B. (2020). The role of decision support systems in attenuating racial biases in healthcare delivery. *Management science*, *66*(11), 5171-5181.

Greenwood, B. N., Hardeman, R. R., Huang, L., & Sojourner, A. (2020). Physician–patient racial concordance and disparities in birthing mortality for newborns. *Proceedings of the National Academy of Sciences*, *117*(35), 21194-21200.

Hannun, A. Y., Rajpurkar, P., Haghpanahi, M., Tison, G. H., Bourn, C., Turakhia, M. P., & Ng, A. Y. (2019). Cardiologist-level arrhythmia detection and classification in ambulatory electrocardiograms using a deep neural network. *Nature medicine*, *25*(1), 65-69.

Himmelstein, G., Bates, D., & Zhou, L. (2022). Examination of stigmatizing language in the electronic health record. *JAMA network open*, *5*(1), e2144967-e2144967.

Holroyd, J., Sweetman, J., Brownstein, M., & Saul, J. (2016). The heterogeneity of implicit bias. *Implicit bias and philosophy*, *1*, 80-103.

Huang, K., Altosaar, J., & Ranganath, R. (2019). Clinicalbert: Modeling clinical notes and predicting hospital readmission. *arXiv preprint arXiv:1904.05342*.

Hugot, J. P., Chamaillard, M., Zouali, H., Lesage, S., Cézard, J. P., Belaiche, J., ... & Thomas, G. (2001). Association of NOD2 leucine-rich repeat variants with susceptibility to Crohn's disease. *Nature*, *411*(6837), 599-603.

Jia, R., & Liang, P. (2017, September). Adversarial Examples for Evaluating Reading Comprehension Systems. In *Proceedings of the 2017 Conference on Empirical Methods in Natural Language Processing* (pp. 2021-2031).

Johnson, A. E., Pollard, T. J., Shen, L., Lehman, L. W. H., Feng, M., Ghassemi, M., ... & Mark, R. G. (2016). MIMIC-III, a freely accessible critical care database. *Scientific data*, *3*(1), 1-9.

Kallus, N., Mao, X., & Zhou, A. (2022). Assessing algorithmic fairness with unobserved protected class using data combination. *Management Science*, *68*(3), 1959-1981.

Kelly, J. F., Wakeman, S. E., & Saitz, R. (2015). Stop talking 'dirty': clinicians, language, and quality of care for the leading cause of preventable death in the United States. *The American journal of medicine*, *128*(1), 8-9.

Kung, T. H., Cheatham, M., Medenilla, A., Sillos, C., De Leon, L., Elepaño, C., ... & Tseng, V. (2023). Performance of ChatGPT on USMLE: Potential for AI-assisted medical education using large language models. *PLOS Digital Health*, *2*(2), e0000198.

Lamont, E. B., & Christakis, N. A. (2003). Complexities in prognostication in advanced cancer: to help them live their lives the way they want to. *Jama*, *290*(1), 98-104.

Landi, H. (2021). Healthcare AI investment will shift to these 5 areas in the next 2 years: survey. *Fierce Healthcare* (Mar 9), https://www.fiercehealthcare.com/tech/healthcare-executives-want-ai-adoption-to-ramp-up-here-s-5-areas-they-plan-to-focus-future.

Lee, J., Yoon, W., Kim, S., Kim, D., Kim, S., So, C. H., & Kang, J. (2020). BioBERT: a pre-trained biomedical language representation model for biomedical text mining. *Bioinformatics*, *36*(4), 1234-1240.

Li, J., Monroe, W., & Jurafsky, D. (2016). Understanding neural networks through representation erasure. *arXiv preprint arXiv:1612.08220*.

Li, Y., Rao, S., Solares, J. R. A., Hassaine, A., Ramakrishnan, R., Canoy, D., ... & Salimi-Khorshidi, G. (2020). BEHRT: transformer for electronic health records. *Scientific reports*, *10*(1), 1-12.

Lin, Y. K., Lin, M., & Chen, H. (2019). Do electronic health records affect quality of care? Evidence from the HITECH Act. *Information Systems Research*, *30*(1), 306-318.

Lou, B., & Wu, L. (2021). AI on Drugs: Can Artificial Intelligence Accelerate Drug Development? Evidence from a Large-Scale Examination of Bio-Pharma Firms. *Management Information Systems Quarterly*, *45*(3), 1451-1482.

Malik, N., & Singh, P. V. (2019). Deep learning in computer vision: Methods, interpretation, causation, and fairness. In *Operations Research & Management Science in the Age of Analytics* (pp. 73-100). INFORMS.

Martens, D., & Provost, F. (2014). Explaining data-driven document classifications. *MIS quarterly*, *38*(1), 73-100.

Martin, S., Brown, W. M., Klavans, R., & Boyack, K. W. (2011, January). OpenOrd: an open-source toolbox for large graph layout. In *Visualization and Data Analysis 2011* (Vol. 7868, pp. 45-55). SPIE.

Martin, A. E., D'Agostino, J. A., Passarella, M., & Lorch, S. A. (2016). Racial differences in parental satisfaction with neonatal intensive care unit nursing care. *Journal of Perinatology*, *36*(11), 1001-1007.

McGarvey, L. P., John, M., Anderson, J. A., Zvarich, M., & Wise, R. A. (2007). Ascertainment of cause-specific mortality in COPD: operations of the TORCH Clinical Endpoint Committee. *Thorax*, *62*(5), 411-415.

Mehrabi, N., Morstatter, F., Saxena, N., Lerman, K., & Galstyan, A. (2021). A survey on bias and fairness in machine learning. *ACM Computing Surveys (CSUR)*, *54*(6), 1-35.

Merryweather-Clarke, A. T., Pointon, J. J., Jouanolle, A. M., Rochette, J., & Robson, K. J. (2000). Geography of HFE C282Y and H63D mutations. *Genetic testing*, *4*(2), 183-198.

Miotto, R., Li, L., Kidd, B. A., & Dudley, J. T. (2016). Deep patient: an unsupervised representation to predict the future of patients from the electronic health records. *Scientific reports*, *6*(1), 1-10.

Mountain, J. L., & Cavalli-Sforza, L. L. (1997). Multilocus genotypes, a tree of individuals, and human evolutionary history. *The American Journal of Human Genetics*, *61*(3), 705-718.

National Institute on Drug Abuse. (2021). Words Matter–Terms to Use and Avoid When Talking About Addiction.

Negro-Calduch, E., Azzopardi-Muscat, N., Krishnamurthy, R. S., & Novillo-Ortiz, D. (2021). Technological progress in electronic health record system optimization: Systematic review of systematic literature reviews. *International journal of medical informatics*, *152*, 104507.

Obermeyer, Z., Powers, B., Vogeli, C., & Mullainathan, S. (2019). Dissecting racial bias in an algorithm used to manage the health of populations. *Science*, *366*(6464), 447-453.

Opsahl, T., Agneessens, F., & Skvoretz, J. (2010). Node centrality in weighted networks: Generalizing degree and shortest paths. *Social networks*, *32*(3), 245-251.

P Goddu, A., O'Conor, K. J., Lanzkron, S., Saheed, M. O., Saha, S., Peek, M. E., ... & Beach, M. C. (2018). Do words matter? Stigmatizing language and the transmission of bias in the medical record. *Journal of general internal medicine*, *33*(5), 685-691.

Parikh, R. B., Teeple, S., & Navathe, A. S. (2019). Addressing bias in artificial intelligence in health care. *Jama*, *322*(24), 2377-2378.

Park, J., Saha, S., Chee, B., Taylor, J., & Beach, M. C. (2021a). Physician use of stigmatizing language in patient medical records. *JAMA Network Open*, *4*(7), e2117052-e2117052.

Park, N., Jang, K., Cho, S., & Choi, J. (2021b). Use of offensive language in human-artificial intelligence chatbot interaction: The effects of ethical ideology, social competence, and perceived humanlikeness. *Computers in Human Behavior*, *121*, 106795.

Patel, S. B., & Lam, K. (2023). ChatGPT: the future of discharge summaries?. *The Lancet Digital Health*, *5*(3), e107-e108.

Paulus, J. K., & Kent, D. M. (2020). Predictably unequal: understanding and addressing concerns that algorithmic clinical prediction may increase health disparities. *NPJ digital medicine*, *3*(1), 99.

Payne, B. K., Vuletich, H. A., & Lundberg, K. B. (2017). The bias of crowds: How implicit bias bridges personal and systemic prejudice. *Psychological Inquiry*, *28*(4), 233-248.

Perrigo, B. (2023). OpenAI Used Kenyan Workers on Less Than $2 Per Hour to Make ChatGPT Less Toxic. *Time* (Jan 18), https://time.com/6247678/openai-chatgpt-kenya-workers/.

Posner, T., & Fei-Fei, L. (2020). AI will change the world, so it's time to change AI. *Nature*, *588*(7837), S118-S118.

Rajpurkar, P., Chen, E., Banerjee, O., & Topol, E. J. (2022). AI in health and medicine. *Nature medicine*, *28*(1), 31-38.

Ridker, P. M., Miletich, J. P., Hennekens, C. H., & Buring, J. E. (1997). Ethnic distribution of factor V Leiden in 4047 men and women: implications for venous thromboembolism screening. *Jama*, *277*(16), 1305-1307.

Risch, N., Burchard, E., Ziv, E., & Tang, H. (2002). Categorization of humans in biomedical research: genes, race and disease. *Genome biology*, *3*(7), 1-12.

Rosen, M. A., DiazGranados, D., Dietz, A. S., Benishek, L. E., Thompson, D., Pronovost, P. J., & Weaver, S. J. (2018). Teamwork in healthcare: Key discoveries enabling safer, high-quality care. *American Psychologist*, *73*(4), 433.

Rosenberg, N. A., Pritchard, J. K., Weber, J. L., Cann, H. M., Kidd, K. K., Zhivotovsky, L. A., & Feldman, M. W. (2002). Genetic structure of human populations. *science*, *298*(5602), 2381-2385.

Rudin, C. (2019). Stop explaining black box machine learning models for high stakes decisions and use interpretable models instead. *Nature machine intelligence*, *1*(5), 206-215.

Samtani, S., Zhu, H., Padmanabhan, B., Chai, Y., Chen, H., & Nunamaker Jr, J. F. (2023). Deep learning for information systems research. *Journal of Management Information Systems*, *40*(1), 271-301.

Sarzynska-Wawer, J., Wawer, A., Pawlak, A., Szymanowska, J., Stefaniak, I., Jarkiewicz, M., & Okruszek, L. (2021). Detecting formal thought disorder by deep contextualized word representations. *Psychiatry Research*, *304*, 114135.

Hollister, S. (2023). Google denies Bard was trained with ChatGPT data. *The Verge* (Mar 29), https://www.theverge.com/2023/3/29/23662621/google-bard-chatgpt-sharegpt-training-denies.

Schönberger, D. (2019). Artificial intelligence in healthcare: a critical analysis of the legal and ethical implications. *International Journal of Law and Information Technology*, *27*(2), 171-203.

Schwartz, R., Vassilev, A., Greene, K., Perine, L., Burt, A., & Hall, P. (2022). Towards a standard for identifying and managing bias in artificial intelligence. *NIST Special Publication*, *1270*, 1-77.

Seinen, T. M., Fridgeirsson, E. A., Ioannou, S., Jeannetot, D., John, L. H., Kors, J. A., ... & Rijnbeek, P. R. (2022). Use of unstructured text in prognostic clinical prediction models: a systematic review. *Journal of the American Medical Informatics Association*, *29*(7), 1292-1302.

Shen, M. C., Lin, J. S., & Tsay, W. (1997). High prevalence of antithrombin III, protein C and protein S deficiency, but no factor V Leiden mutation in venous thrombophilic Chinese patients in Taiwan. *Thrombosis research*, *87*(4), 377-385.

Sparrowe, R. T., Liden, R. C., Wayne, S. J., & Kraimer, M. L. (2001). Social networks and the performance of individuals and groups. *Academy of management journal*, *44*(2), 316-325.

Stephens, J. C., Reich, D. E., Goldstein, D. B., Shin, H. D., Smith, M. W., Carrington, M., ... & Dean, M. (1998). Dating the origin of the CCR5-Δ32 AIDS-resistance allele by the coalescence of haplotypes. *The American Journal of Human Genetics*, *62*(6), 1507-1515.

Stephens, J. C., Schneider, J. A., Tanguay, D. A., Choi, J., Acharya, T., Stanley, S. E., ... & Vovis, G. F. (2001). Haplotype variation and linkage disequilibrium in 313 human genes. *Science*, *293*(5529), 489-493.

Stokel-Walker, C. (2023). ChatGPT listed as author on research papers: many scientists disapprove. *Nature* (Jan 18). https://www.nature.com/articles/d41586-023-00107-z.

Sun, M., Oliwa, T., Peek, M. E., & Tung, E. L. (2022). Negative Patient Descriptors: Documenting Racial Bias In The Electronic Health Record: Study examines racial bias in the patient descriptors used in the electronic health record. *Health Affairs*, *41*(2), 203-211.

Tamayo-Sarver, J. H., Hinze, S. W., Cydulka, R. K., & Baker, D. W. (2003). Racial and ethnic disparities in emergency department analgesic prescription. *American journal of public health*, *93*(12), 2067-2073.

The White House. (2022). A blueprint for an AI bill of rights. Report, The Office of Science and Technology Policy (OSTP), The Executive Office of the President, Washington, DC.

Vaswani, A., Shazeer, N., Parmar, N., Uszkoreit, J., Jones, L., Gomez, A. N., ... & Polosukhin, I. (2017). Attention is all you need. *Advances in neural information processing systems*, *30*.

Vingerhoets, A. J., Bylsma, L. M., & De Vlam, C. (2013). Swearing: A biopsychosocial perspective. *Psihologijske teme*, *22*(2), 287-304.

Vyas, D. A., Eisenstein, L. G., & Jones, D. S. (2020). Hidden in plain sight—reconsidering the use of race correction in clinical algorithms. *New England Journal of Medicine*, *383*(9), 874-882.

Wang, S., McDermott, M. B., Chauhan, G., Ghassemi, M., Hughes, M. C., & Naumann, T. (2020, April). Mimic-extract: A data extraction, preprocessing, and representation pipeline for mimic-iii. In *Proceedings of the ACM conference on health, inference, and learning* (pp. 222-235).

Werder, K., Curtis, A., Reynolds, S., & Satterfield, J. (2022). Addressing bias and stigma in the language we use with persons with opioid use disorder: A narrative review. *Journal of the American Psychiatric Nurses Association*, *28*(1), 9-22.

Wen, Z., Lu, X. H., & Reddy, S. (2020, November). MeDAL: Medical Abbreviation Disambiguation Dataset for Natural Language Understanding Pretraining. In *Proceedings of the 3rd Clinical Natural Language Processing Workshop* (pp. 130-135).

Wilson, J. F., Weale, M. E., Smith, A. C., Gratrix, F., Fletcher, B., Thomas, M. G., ... & Goldstein, D. B. (2001). Population genetic structure of variable drug response. *Nature genetics*, *29*(3), 265-269.

World Health Organization. (2021). Ethics and governance of artificial intelligence for health: WHO guidance.

Xiao, C., Choi, E., & Sun, J. (2018). Opportunities and challenges in developing deep learning models using electronic health records data: a systematic review. *Journal of the American Medical Informatics Association*, *25*(10), 1419-1428.

Yamazaki, K., Takazoe, M., Tanaka, T., Kazumori, T., & Nakamura, Y. (2002). Absence of mutation in the NOD2/CARD15 gene among 483 Japanese patients with Crohn's disease. *Journal of human genetics*, *47*(9), 469-472.

Yang, D., & Song, J. (2010, October). Web content information extraction approach based on removing noise and content-features. In *2010 International conference on web information systems and mining* (Vol. 1, pp. 246-249). IEEE.

Yu, S., Chai, Y., Chen, H., Sherman, S., & Brown, R. (2021). Wearable Sensor-based Chronic Condition Severity Assessment: An Adversarial Attention-based Deep Multisource Multitask Learning Approach. *Forthcoming in MIS Quarterly*.

Yu, M. C., Skipper, P. L., Taghizadeh, K., Tannenbaum, S. R., Chan, K. K., Henderson, B. E., & Ross, R. K. (1994). Acetylator phenotype, aminobiphenyl-hemoglobin adduct levels, and bladder cancer risk in white, black, and Asian men in Los Angeles, California. *JNCI: Journal of the National Cancer Institute*, *86*(9), 712-716.

Yue, X., Nouiehed, M., & Al Kontar, R. (2022). Gifair-fl: A framework for group and individual fairness in federated learning. *INFORMS Journal on Data Science*.

Zhang, H., Lu, A. X., Abdalla, M., McDermott, M., & Ghassemi, M. (2020, April). Hurtful words: quantifying biases in clinical contextual word embeddings. In *proceedings of the ACM Conference on Health, Inference, and Learning* (pp. 110-120).

Zhang, W., & Ram, S. (2020). A Comprehensive Analysis of Triggers and Risk Factors for Asthma Based on Machine Learning and Large Heterogeneous Data Sources. *MIS Quarterly*, *44*(1).

Zhou, C., Li, Q., Li, C., Yu, J., Liu, Y., Wang, G., ... & Sun, L. (2023). A comprehensive survey on pretrained foundation models: A history from bert to chatgpt. *arXiv preprint arXiv:2302.09419*.

# Online Appendix A: Literature Review on Explainable AI (XAI)

We review the Explainable AI (XAI) literature to guide us on how to examine the effect of SL on AI performance and fairness systematically. Explainable AI is a rapidly evolving field which aims to develop AI models that are transparent, trustworthy, and can be easily understood by humans (Gunning et al. 2019). The emergence of XAI is driven by the fact that many AI models are a "black box" that cannot convince users to entrust the models and make crucial decisions accordingly (Deeks 2019, Wang et al. 2019). Due to the high demand for transparency and trust in clinical prediction, XAI has received increasing attention in the healthcare field (Yang 2022, Lauritsen et al. 2020). In the IS field, studies have been increasingly incorporating explainability into the model design (Zhou et al. 2023, Xie et al. 2022).

XAI methods can be broadly categorized into model-specific and model-agnostic methods (Frye et al. 2020). Model-specific methods aim to make an AI model more interpretable, while model-agnostic methods can be applied to any black-box model. Model agnosticism is often considered a desirable property in XAI, as it allows for a wider range of models to be explained and makes the explanation technique more versatile (Hashemi 2023). This can be especially important in situations such as clinical prediction where a general model may not exist and the model's inner states are inaccessible.

Regarding NLP models, XAI methods proposed by prior research can be categorized into six directions (Wallace et al. 2020): 1) probing internal representations, 2) testing model behavior using challenge and diagnostic sets, 3) baking interpretability into the model, 4) looking for global decision rules, 5) looking at input features, and 6) looking at training examples.

The concept of probing internal representations involves examining a model's hidden layers or internal representations to gain insight into how it makes decisions (Gunning and Aha 2019). To assess a model's weaknesses and limitations, researchers can use challenge sets to determine how it performs under adverse conditions, while diagnostic sets are designed to test its ability to recognize different types of features or relationships in the data (D'Amour et al. 2022). Baking interpretability into a model involves designing the model architecture and training process to promote transparency and explainability, although this is typically only applicable to simple models (Das and Rad 2020). When looking for global decision rules, researchers aim to identify patterns and rules (e.g., a trigger word) that can explain the overall behavior of a model across datasets (Ribeiro et al. 2018). While these directions have been widely used in the literature, their focus is not on the inputs, which may not serve our purpose of examining the impact of SL in EHR notes.

In contrast, looking at input features and looking at training examples are two viable directions. Looking at input features is motivated by the following question: *which part of the inputs are responsible for the prediction?* Figure A1 illustrates a general process of looking at input features. It involves identifying which features are most important or influential in the model's predictions. There are two classes of methods for looking at the input feature direction. The first class is the saliency maps, a popular method in XAI that aims to highlight the importance of input features with visualizations. The most common saliency map is the visualization of attention scores. The attention scores indicate how much attention the model is giving to each input feature. These scores can be visualized as a heat map, where the input features with higher scores are represented with brighter colors, making it easier for humans to interpret the model's behavior (Mohseni et al. 2021). Three other common methods of saliency maps are layer relevance propagation (Montavon et al. 2019), the leave-one-out strategy (Li et al. 2016), and Local Interpretable Model-agnostic Explanations (LIME) and SHapley Additive exPlanations (SHAP) (Lundberg and Lee 2017, Ribeiro et al. 2016). Saliency maps are intuitive, scalable, and model-agnostic; meanwhile, they do not require any modifications to the model architecture or training process, making them a non-intrusive method for analyzing model behavior.

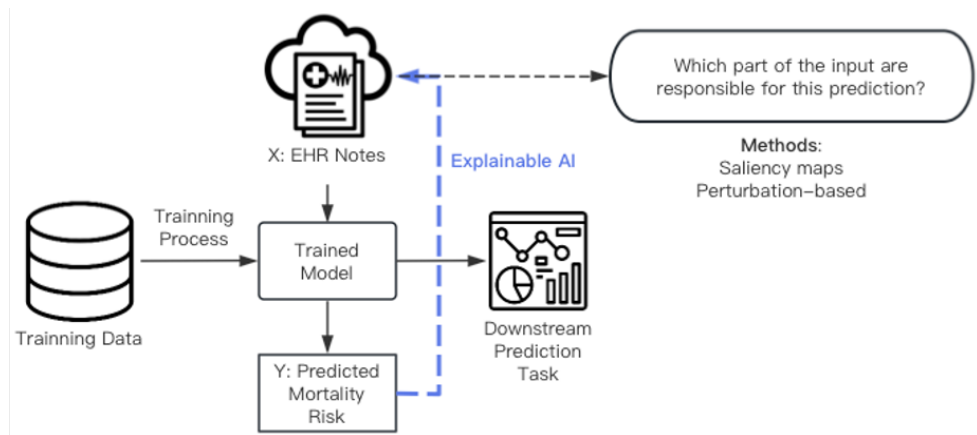

**Figure A1.** A general process of looking at input features

Another class is the perturbation-based methods which investigate the model properties by perturbing the inputs and observing the changes in the outputs (Ivanovs et al. 2021). Compared with saliency maps that attribute the importance of input features without directly altering the inputs, perturbation-based methods systematically alter the input data in various ways to understand how the model responds to changes in the input. Input reduction and adversarial perturbations are two common perturbation-based approaches. Input reduction iteratively removes the least important word from the inputs to observe whether models have overconfidence and oversensitivity, which serves as a robustness check to saliency map approaches (Feng et al. 2018). Adversarial perturbations aim to add adversarial examples that could change the prediction results, thereby understanding the model's weakness and explainability (Ebrahimi et al. 2018, Iyyer et al. 2018, Ribeiro et al. 2018). For instance, Jia and Liang (2017) investigate how a question-answering model can be misled by adding adversarial sentences. Looking at input features are usually model-agnostic, easy to compute, and faithful to underlying models, and therefore, they are suitable for examining whether SL affects AI performance and fairness.

Although understanding a model's performance on new, unseen data is essential, analyzing the training examples can also provide valuable insights into how the model learns and makes predictions (Koh and Liang 2017). Looking at training examples is motivated by two questions: *which training examples were responsible for this prediction? Which examples, if removed, would change the loss a lot?* Figure A2 illustrates a general process of looking at training examples. Specifically, examining input features in the training set can help identify which features are most important for the model to classify data correctly and for identifying potential biases in the data or model (Han et al. 2020). This information can be used to improve the data preprocessing stage as well as the fairness and accuracy of the model. Data influence is a representative method that aims to identify the most influential samples or input features in the training data (Koh and Liang 2017). Conversely, it can be used to identify the least important features or features that are responsible for AI biases, thereby promoting AI fairness. If specific features in the training data are found responsible for erroneous or biased predictions, a quick fix is to remove them from the training data (Yang and Song 2010). However, any features may contain information value, and removing them blindly may cause information loss (Al Sharou et al. 2021, Cordeiro and Brazdil 2004).

Whether removing SL from EHR notes at the training step can help improve AI performance and fairness has not been examined.

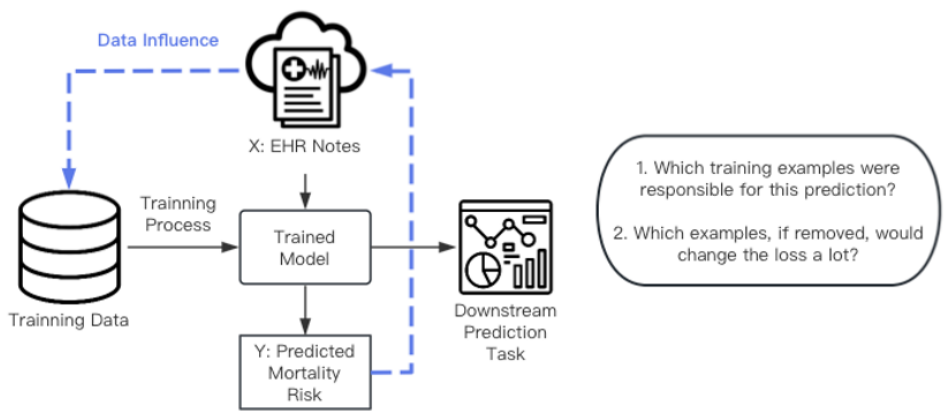

**Figure A2.** A general process of looking at training examples

## Online Appendix B: Technical Details of Experiments

We follow previous studies to conduct data preprocessing. The first preprocessing step is data cleaning. Specifically, since our aim is to predict the mortality of ICU patients using the first 24 hours of ICU stays as the observational window, we only include EHR notes of patients who survived the first 24 hours and stayed longer than 24 hours. Also, we exclude the discharge notes since this type of notes means that the patients were not deceased and thus is not necessary for prediction. In addition, we remove the notes that contain less than 20 words (less than 1% of the average length, 2,058 words) since they are mostly noises or have almost no informative content for mortality prediction.

Second, EHR notes are often excessively lengthy, which becomes a problem as many advanced NLP models struggle with long inputs due to the difficulty of capturing long-range dependencies. For transformer-based models, a critical issue with long inputs is the computational complexity of processing a large number of tokens in a sequence. To this end, we conduct segmentation, breaking the input text into smaller pieces or segments, which can then be processed by the model in parallel or sequentially. The last preprocessing step is tokenization, which is a critical step in NLP. It processes and analyzes text by breaking it down into smaller and structured pieces to make it understandable for AI models.

It is worth noting that after asking the opinions from domain experts, we excluded seven keywords from the SL list identified by Himmelstein et al. (2022): *fail, fails, failed, failure, control, controls, controlled*. In the ICU setting, These words are often related to the patients' conditions (e.g., cardiac failure, glycemic control) instead of SL. We utilize regular expressions to identify whether each note contains the SL keywords in the list, and if yes, we label it as 1, otherwise 0, as a label indicating if an EHR note is SL-related.

We train and test a Transformer model in this study. The core of Transformer is the self-attention mechanism, which aims to relate each word with every other word of the sentence (Vaswani et al. 2017). To compute self-attention, we need to create three vectors named query (Q), key (K), and value (V), which are calculated by matrix multiplication between the input features (e.g., representations of an EHR note) and three learnable weight matrices. Then, query, key, and value vectors are used to perform the following calculation:

$$Self\,Attention = softmax\left(\frac{QK^T}{\sqrt{d_K}}\right)V \tag{1}$$

The output is computed as a weighted $V$, where weights are normalized alignment scores calculated based on the relationship between $Q$ and $K$.

## Online Appendix C: Additional Data Description and Analyses

To understand SL is more related to what kinds of clinicians, we first investigate the SL by note category, as shown in Table A1. Consultation has the highest percentage of SL with 32.26%, while it only has 62 notes in the data and is a rare category. 25.83% of Physician notes have SL, which is the second highest in the data, and it represents a major type of note category. In contrast, nursing notes are the most common note category, while they have a relatively low percentage of SL (10.59% for nursing and 4.27% for nursing/other). These results suggest that different types of clinicians may have different behavior in writing SL, and physicians might be a major source of SL in EHR notes.

**Table A1**. Number of EHR notes by SL and category

| Note Category | Total # of Notes | Stig | # of Notes | % of Stig Notes |
|---|---|---|---|---|
| Case Management | 344 | No | 332 | 96.51% |
|  |  | Yes | 12 | 3.49% |
| Consult | 62 | No | 42 | 67.74% |
|  |  | Yes | 20 | 32.26% |
| General | 3,664 | No | 3,380 | 92.25% |
|  |  | Yes | 284 | 7.75% |
| Nursing | 105,931 | No | 94,717 | 89.41% |
|  |  | Yes | 11,214 | 10.59% |
| Nursing/other | 171,680 | No | 164,344 | 95.73% |
|  |  | Yes | 7,336 | 4.27% |
| Nutrition | 3,586 | No | 3,349 | 93.39% |
|  |  | Yes | 237 | 6.61% |
| Pharmacy | 32 | No | 32 | 100.00% |
|  |  | Yes | 0 | 0.00% |
| Physician | 63,531 | No | 47,119 | 74.17% |
|  |  | Yes | 16,412 | 25.83% |
| Radiology | 203,913 | No | 201,755 | 98.94% |
|  |  | Yes | 2,158 | 1.06% |
| Rehab Services | 2,644 | No | 2,446 | 92.51% |
|  |  | Yes | 198 | 7.49% |
| Respiratory | 9,939 | No | 9,845 | 99.05% |
|  |  | Yes | 94 | 0.95% |
| Social Work | 1,271 | No | 1,038 | 81.67% |
|  |  | Yes | 233 | 18.33% |

As shown in Figure A3, most SL used by clinicians is related to the substance use disorder (SUD) of patients, which is considered SL by prior studies and the National Institutes of Health (NIH).[7]

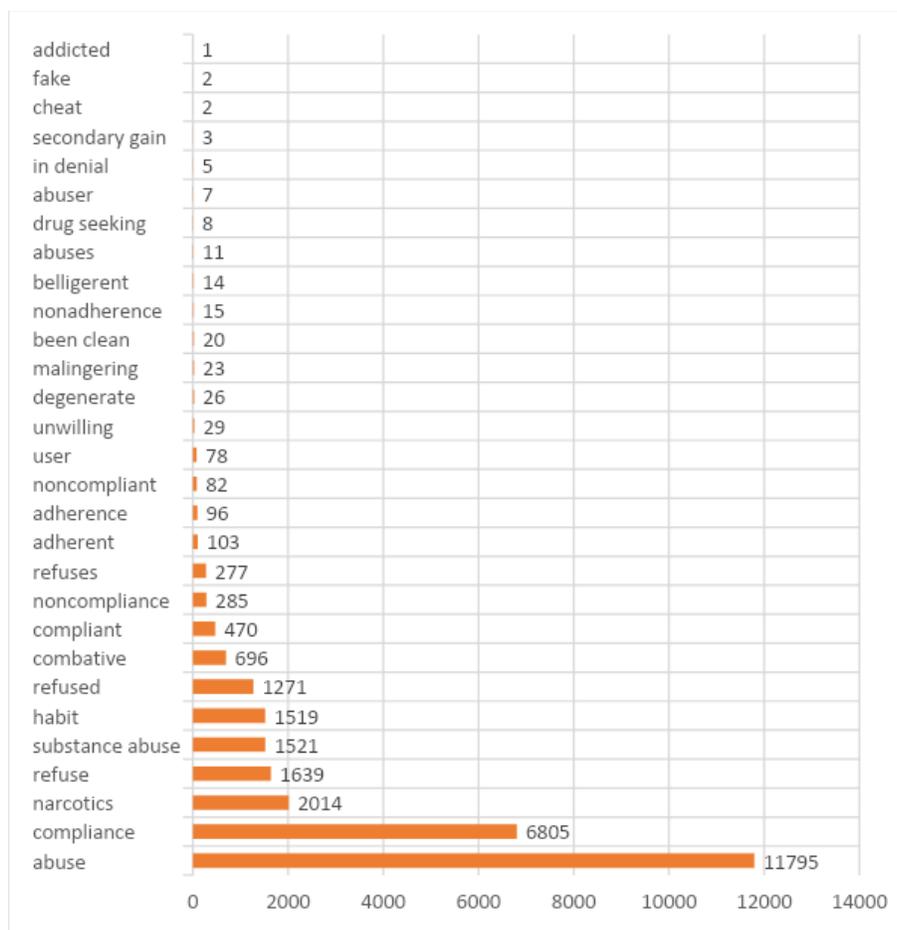

**Figure A3.** Term frequency of SL in EHR notes

Our main results provide convincing evidence that removing SL can significantly improve AI performance. To obtain a further understanding of the results, we drill down to the underlying mechanisms by focusing on the attention scores. As shown in Fig A4, SL can attract and dilute the attention of AI models. On average, EHR in our data has 2,058 words, and the attention for each word is 4.86E-04. In contrast, there are 2.29 SL words in EHR notes on average, and the attention score on an SL word is 3.08E-03. In other words, SL in EHR attracted over 6 times more attention than average. Quantifying the attention scores on SL and other words in EHR notes allows us to demonstrate the dilution effect of SL on AI performance, thereby partially explaining why removing SL, a specific type of noise in EHR notes, can improve AI performance.

| **Example 1** | | | | | | | | | |
|---|---|---|---|---|---|---|---|---|---|
| Pt | asking | for | clothes, | insisting | she | wanted | to | get | dressed. |
| 0.0003 | 0.0362 | 0.0002 | 0.0857 | 0.4412 | 0.0015 | 0.0860 | 0.0001 | 0.0230 | 0.1153 |

| Tried | to | explain | that | she | is | in | the | hospital | but | she | would | have | none | of | that. |
|---|---|---|---|---|---|---|---|---|---|---|---|---|---|---|---|
| 0.0769 | 0.0001 | 0.0859 | 0.0030 | 0.0015 | 0.0003 | 0.0002 | 0.0001 | 0.0118 | 0.0013 | 0.0015 | 0.0111 | 0.0028 | 0.0111 | 0.0001 | 0.0030 |

| **Example 2** | | | | | | | | | | | | | | |
|---|---|---|---|---|---|---|---|---|---|---|---|---|---|---|
| Unclear | if | L | sided | weakness | is | worse | as | pt | can't | cooperate | w/ | full | neuro | exam. |
| 0.0397 | 0.0074 | 0.0015 | 0.0350 | 0.0656 | 0.0006 | 0.1159 | 0.0010 | 0.0006 | 0.3149 | 0.3839 | 0.0021 | 0.0103 | 0.0080 | 0.0138 |

**Figure A4**. Examples of attention scores on SL in EHR notes

In the clinician collaborative network, the nodes are clinicians, and there is an edge if two clinicians have served the same patient. Since the data comprises many years of EHR notes of only 1,492 clinicians, there is a high chance that two clinicians have served several patients together before, but that does not necessarily mean they worked closely as a team. To identify central clinicians who worked very closely with each other, we give the edge weights using the number of patients that two clinicians have served in the data. The highest edge weight is 3,746, the lowest is 1, the average is 35.05, and the standard deviation is 86.50. The high standard deviation suggests that some clinicians work highly closely with their peers, and they have higher centrality in the network (i.e., central clinicians), while others have relatively low centrality (non-central clinicians). In a weighted network, the concept of degree centrality is extended to account for the weights of the edges, and a common measure is known as node strength, which is calculated by summing the weights of all the edges a node has (Opsahl et al. 2010). Thus, we categorize the nodes into central and non-central ones based on their node strength. This method allows us to investigate the association between the centrality of clinicians (i.e., familiarity with their peers) and SL writing behavior. The complete collaborative network of clinicians is illustrated in Figure A5.

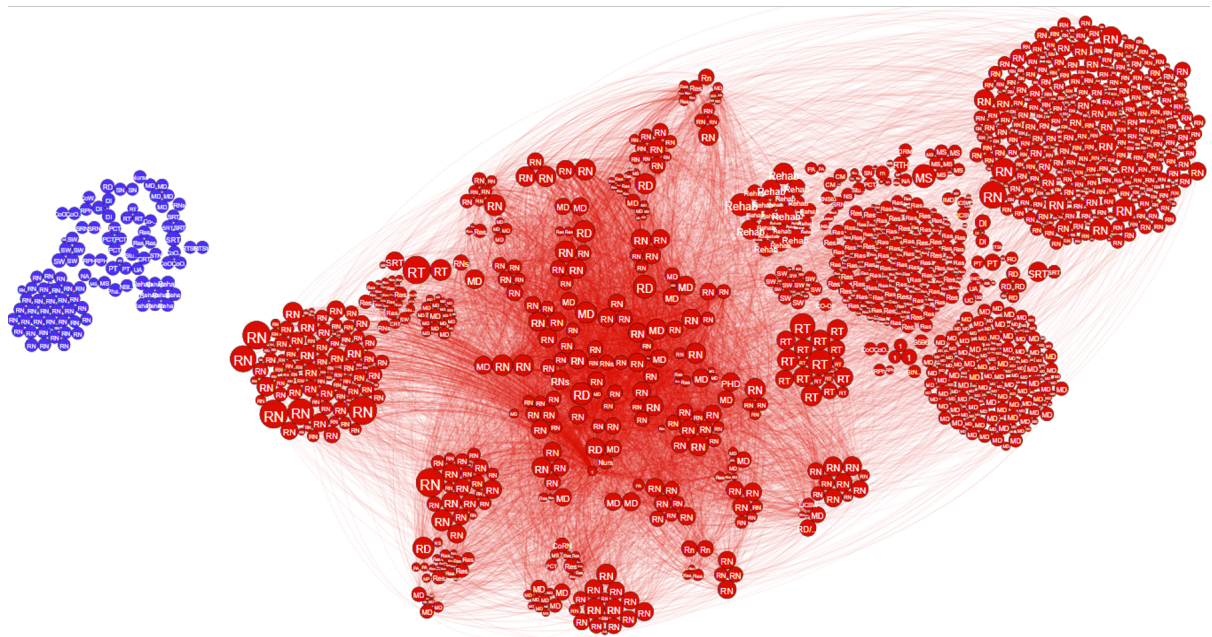

**Figure A5**. Complete collaborative network of clinicians. **Note:** clinicians who have written SL in the EHR notes are in red, and those who have never written SL are in blue. Each edge suggests that two clinicians served the same patient.

# References

Al Sharou, K., Li, Z., & Specia, L. (2021, September). Towards a better understanding of noise in natural language processing. In *Proceedings of the International Conference on Recent Advances in Natural Language Processing (RANLP 2021)* (pp. 53-62).

Cordeiro, J., & Brazdil, P. (2004). Learning Text Extraction Rules, without Ignoring Stop Words. In *PRIS* (pp. 128-138).

D'Amour, A., Heller, K., Moldovan, D., Adlam, B., Alipanahi, B., Beutel, A., ... & Sculley, D. (2022). Underspecification presents challenges for credibility in modern machine learning. *The Journal of Machine Learning Research*, *23*(1), 10237-10297.

Das, A., & Rad, P. (2020). Opportunities and challenges in explainable artificial intelligence (xai): A survey. *arXiv preprint arXiv:2006.11371*.

Deeks, A. (2019). The judicial demand for explainable artificial intelligence. *Columbia Law Review*, *119*(7), 1829-1850.

Ebrahimi, J., Lowd, D., & Dou, D. (2018). On adversarial examples for character-level neural machine translation. *arXiv preprint arXiv:1806.09030*.

Feng, S., Wallace, E., Grissom II, A., Iyyer, M., Rodriguez, P., & Boyd-Graber, J. (2018). Pathologies of Neural Models Make Interpretations Difficult. In *Proceedings of the 2018 Conference on Empirical Methods in Natural Language Processing* (pp. 3719-3728).

Fernandez, C., Provost, F., & Han, X. (2022). Explaining data-driven decisions made by AI systems: the counterfactual approach. *MIS Quarterly*, *46*(3).

Frye, C., Rowat, C., & Feige, I. (2020). Asymmetric shapley values: incorporating causal knowledge into model-agnostic explainability. *Advances in Neural Information Processing Systems*, *33*, 1229-1239.

Gunning, D., & Aha, D. (2019). DARPA's explainable artificial intelligence (XAI) program. *AI magazine*, *40*(2), 44-58.

Gunning, D., Stefik, M., Choi, J., Miller, T., Stumpf, S., & Yang, G. Z. (2019). XAI—Explainable artificial intelligence. *Science robotics*, *4*(37), eaay7120.

Han, S. H., Kwon, M. S., & Choi, H. J. (2020). EXplainable AI (XAI) approach to image captioning. *The Journal of Engineering*, *2020*(13), 589-594.

Hashemi, M. (2023). Who wants what and how: a Mapping Function for Explainable Artificial Intelligence. *arXiv preprint arXiv:2302.03180*.

Himmelstein, G., Bates, D., & Zhou, L. (2022). Examination of stigmatizing language in the electronic health record. *JAMA network open*, *5*(1), e2144967-e2144967.

Ivanovs, M., Kadikis, R., & Ozols, K. (2021). Perturbation-based methods for explaining deep neural networks: A survey. *Pattern Recognition Letters*, *150*, 228-234.

Iyyer, M., Wieting, J., Gimpel, K., & Zettlemoyer, L. (2018). Adversarial Example Generation with Syntactically Controlled Paraphrase Networks. In *Proceedings of NAACL-HLT* (pp. 1875-1885).

Koh, P. W., & Liang, P. (2017, July). Understanding black-box predictions via influence functions. In *International conference on machine learning* (pp. 1885-1894). PMLR.

Lauritsen, S. M., Kristensen, M., Olsen, M. V., Larsen, M. S., Lauritsen, K. M., Jørgensen, M. J., ... & Thiesson, B. (2020). Explainable artificial intelligence model to predict acute critical illness from electronic health records. *Nature communications*, *11*(1), 3852.

Li, J., Monroe, W., & Jurafsky, D. (2016). Understanding neural networks through representation erasure. *arXiv preprint arXiv:1612.08220*.

Lundberg, S. M., & Lee, S. I. (2017). A unified approach to interpreting model predictions. *Advances in neural information processing systems*, *30*.

Mohseni, S., Block, J. E., & Ragan, E. (2021, April). Quantitative evaluation of machine learning explanations: A human-grounded benchmark. In *26th International Conference on Intelligent User Interfaces* (pp. 22-31).

Montavon, G., Binder, A., Lapuschkin, S., Samek, W., & Müller, K. R. (2019). Layer-wise relevance propagation: an overview. *Explainable AI: interpreting, explaining and visualizing deep learning*, 193-209.

Opsahl, T., Agneessens, F., & Skvoretz, J. (2010). Node centrality in weighted networks: Generalizing degree and shortest paths. *Social networks*, *32*(3), 245-251.

Ribeiro, M. T., Singh, S., & Guestrin, C. (2018, April). Anchors: High-precision model-agnostic explanations. In *Proceedings of the AAAI conference on artificial intelligence* (Vol. 32, No. 1).

Ribeiro, M. T., Singh, S., & Guestrin, C. (2016). Model-agnostic interpretability of machine learning. *arXiv preprint arXiv:1606.05386*.

Vaswani, A., Shazeer, N., Parmar, N., Uszkoreit, J., Jones, L., Gomez, A. N., ... & Polosukhin, I. (2017). Attention is all you need. *Advances in neural information processing systems*, *30*.

Xie, J., Chai, Y., & Liu, X. (2022, January). An Interpretable Deep Learning Approach to Understand Health Misinformation Transmission on YouTube. In *Proceedings Of The 55th Hawaii International Conference On System Sciences*.

Wallace, E., Gardner, M., & Singh, S. (2020, November). Interpreting predictions of NLP models. In *Proceedings of the 2020 Conference on Empirical Methods in Natural Language Processing: Tutorial Abstracts* (pp. 20-23).

Wang, D., Yang, Q., Abdul, A., & Lim, B. Y. (2019, May). Designing theory-driven user-centric explainable AI. In *Proceedings of the 2019 CHI conference on human factors in computing systems* (pp. 1-15).

Yang, C. C. (2022). Explainable artificial intelligence for predictive modeling in healthcare. *Journal of healthcare informatics research*, *6*(2), 228-239.

Yang, D., & Song, J. (2010, October). Web content information extraction approach based on removing noise and content-features. In *2010 International conference on web information systems and mining* (Vol. 1, pp. 246-249). IEEE.

Zhou, T., Wang, Y., Yan, L., & Tan, Y. (2023). Spoiled for Choice? Personalized Recommendation for Healthcare Decisions: A Multiarmed Bandit Approach. *Information Systems Research*.